\documentclass[sigconf]{acmart}
\usepackage{multirow}
\usepackage{balance}

\AtBeginDocument{%
  \providecommand\BibTeX{{%
    \normalfont B\kern-0.5em{\scshape i\kern-0.25em b}\kern-0.8em\TeX}}}

\renewcommand\footnotetextcopyrightpermission[1]{} 
\begin{document}
\fancyhead{}
\title{Quality Assessment of Image Super-Resolution: Balancing Deterministic and Statistical Fidelity}



\author{Wei Zhou}
\affiliation{University of Waterloo}
\email{wei.zhou@uwaterloo.ca}
\author{Zhou Wang}
\affiliation{University of Waterloo}
\email{zhou.wang@uwaterloo.ca}


\begin{abstract}
There has been a growing interest in developing image super-resolution (SR) algorithms that convert low-resolution (LR) to higher resolution images, but automatically evaluating the visual quality of super-resolved images remains a challenging problem. Here we look at the problem of SR image quality assessment (SR IQA) in a two-dimensional (2D) space of deterministic fidelity (DF) versus statistical fidelity (SF). This allows us to better understand the advantages and disadvantages of existing SR algorithms, which produce images at different clusters in the 2D space of (DF, SF).
Specifically, we observe an interesting trend from more traditional SR algorithms that are typically inclined to optimize for DF while losing SF, to more recent generative adversarial network (GAN) based approaches that by contrast exhibit strong advantages in achieving high SF but sometimes appear weak at maintaining DF. Furthermore, we propose an uncertainty weighting scheme based on content-dependent sharpness and texture assessment that merges the two fidelity measures into an overall quality prediction named the Super Resolution Image Fidelity (SRIF) index, which demonstrates superior performance against state-of-the-art IQA models when tested on subject-rated datasets.

\end{abstract}



\maketitle

\section{Introduction}

\begin{figure}[t]
	\centering
	\includegraphics[width=9.8cm]{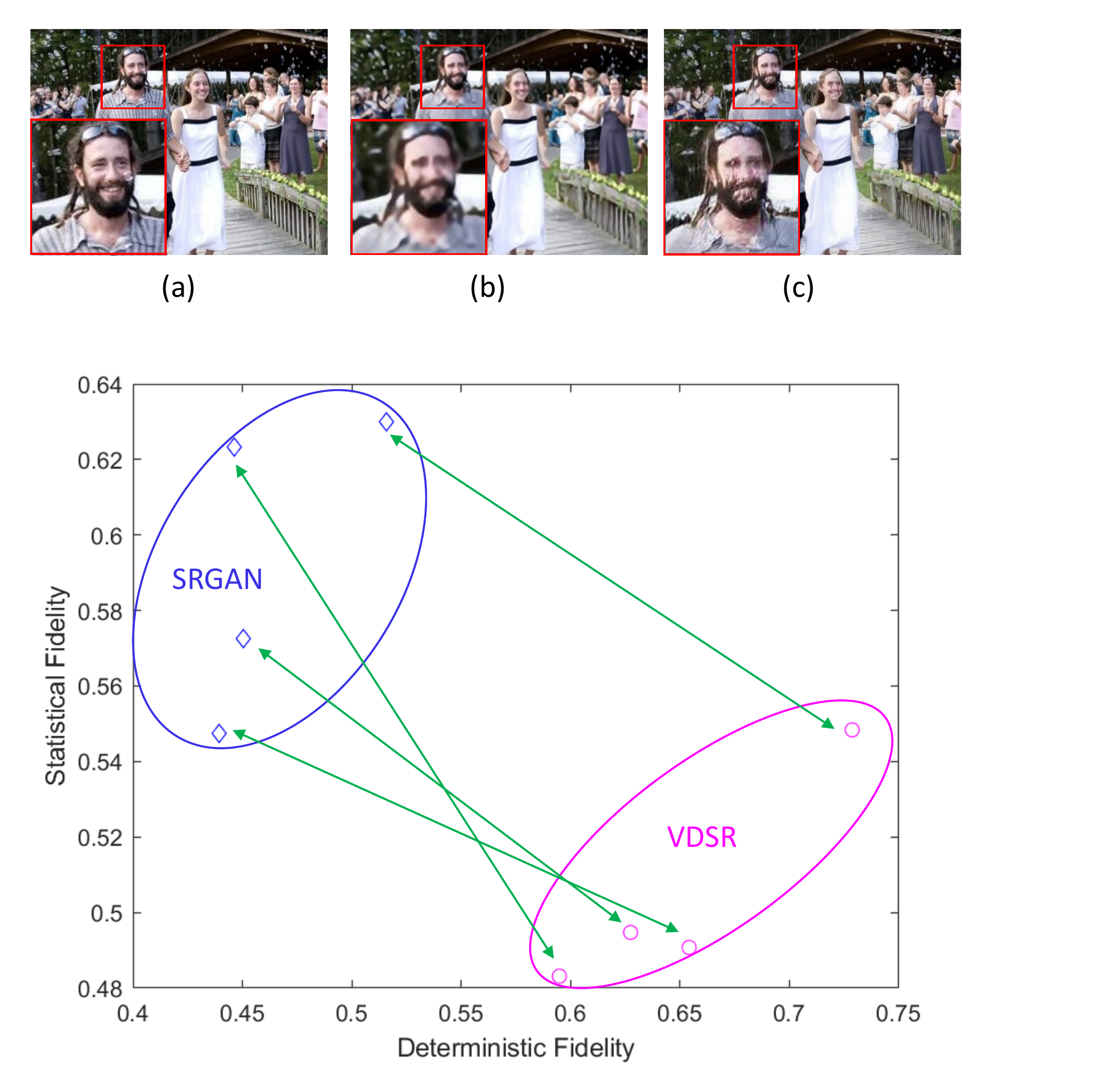}
	\caption{SR images and quality assessment in 2D space of (DF, SF). (a) Original HR image; (b) SR image reconstructed by VDSR \cite{kim2016accurate} at scaling factor 4; (c) SR image reconstructed by SRGAN \cite{ledig2017photo} at scaling factor 4. Bottom: SR images evaluated in (DF, SF) space, where the green arrows link images of the same content. The results reveal drastically different characteristics of SR algorithms.}
	\label{fig1}
\end{figure}

Image super-resolution (SR) targets at ``restoring" the spatial resolution of a given low-resolution (LR) image to a higher quality super-resolved image that potentially recovers the original sharpness and texture details. Image SR is an important and challenging problem that has attracted a great deal of attention in recent years. It has a wide range of real-world applications \cite{park2003super}, including visual communications, video surveillance, medical imaging, and high-definition television, among many others. Image SR is an ill-posed inverse problem because there exist many legitimate SR image solutions for any given single input LR image. Many SR methods have been proposed in the literature, ranging from simple local interpolation algorithms to sophisticated deep learning approaches. With numerous SR methods available, it becomes critically important to automatically assess the perceptual quality of reconstructed SR images for selection and optimization purposes. 
Although there have been a large number of studies in the area of image quality assessment (IQA), few have focused on SR IQA. Subjective quality evaluation is generally regarded as a reliable method~\cite{reibman2006quality,yang2014single,yeganeh2015objective,wang2017perceptual,ma2017learning,shi2019sisrset,zhou2019visual},
but is expensive, time-consuming, and labor-intensive. 
It is also worth noting that certain structural information may be lost or altered during the SR process, but human observers may not be able to notice the existence of such structural loss or change. Therefore, objective image quality or fidelity models that work effectively for SR images are highly desirable.

Nevertheless, existing image quality, fidelity and sharpness models only achieve limited success in SR IQA (as will be shown in Section~\ref{sec:validation}). The challenge is not only in the difficulty in obtaining high correlation with the overall subjective opinions, but perhaps more importantly, on offering meaningful interpretations on the visual artifacts and the intrinsic characteristics of different SR algorithms. An example is shown in Figure ~\ref{fig1}, where the SR images created by a traditional CNN model (VDSR~\cite{kim2016accurate}) and a recent GAN-based model (SRGAN~\cite{ledig2017photo}) are compared with the original reference image. The VDSR generated SR image safely preserves the general structure or deterministic fidelity (DF) according to most existing image quality/fidelity measures such as Peak Signal-to-Noise Ratio (PSNR) and the structural similarity index (SSIM)~\cite{wang2004image}, but loses the fine textures in the original image or the statistical fidelity (SF). By contrast, the SRGAN generated image better preserves the texture complexity/granularity or SF, but meanwhile produces novel structures or artifacts inexistent in the original image, and is thus poor in maintaining the DF. These observations suggest that it may not be appropriate or sufficient to consider SR IQA as a one-dimensional problem. In this work, we develop an objective SR IQA model in a two-dimensional (2D) space of DF versus SF.
As exemplified in the bottom part of Figure ~\ref{fig1}, such a 2D approach clearly characterizes SR algorithms, where images created by different SR algorithms cluster at different regions in the 2D space. Moreover, such a 2D description can be easily merged into a single overall quality/fidelity measure, for which we find an uncertainty weighting scheme leads to superior quality prediction performance against existing approaches when tested using subject-rated datasets.


\section{Related Work}
Early image SR methods are based on direct local interpolations, such as nearest neighbor, bilinear, bicubic, cubic spline interpolation methods~\cite{keys1981cubic}, and orientation-adaptive algorithms~\cite{wang2007new}. Since only very local information is used and certain smoothness constraints are imposed, the reconstructed SR images often appear blurry. Another type of approaches, termed dictionary-based SR~\cite{zhou2019visual} methods, attempt to predict high-resolution (HR) patches from LR ones. These include example-based SR~\cite{freeman2002example}, locally linear embedding~\cite{chang2004super}, sparse representation-based SR~\cite{yang2010image}, semi-coupled dictionary learning model~\cite{wang2012semi}, and consistent coding schemes~\cite{yang2016consistent}. Recently there has been a surge of deep learning-based image SR approaches \cite{wang2020deep}, including convolutional neural network (CNN)-based \cite{guo2020dual,xiao2021self} and generative adversarial network (GAN)-based \cite{ledig2017photo} methods.

Given multiple SR algorithms, one straightforward method to test and compare them is to start with a high-resolution original image, downsample it to an LR image, and then apply the SR algorithms to the LR image to generate multiple SR images. IQA methods can then be employed to assess these images and compare the performance of different SR algorithms. In this scenario, several types of IQA models are applicable, which include full-reference (FR) IQA models that use the original HR image as the pristine reference, no-reference (NR) IQA models that evaluate the generated SR images directly, image sharpness/blurriness models that assess the sharpness/blurriness of the generated SR images, and SR IQA models that are fully dedicated to such SR application scenarios.

The simplest FR methods include PSNR, mean square error (MSE) and mean absolute error (MAE), which have been shown repeatedly in the literature to under-perform the family of structural similarity based models, which include the SSIM index~\cite{wang2004image}, the multi-scale SSIM (MS-SSIM) index~\cite{wang2003multiscale}, the feature similarity (FSIM) index~\cite{zhang2011fsim}, the complex wavelet SSIM (CW-SSIM) index~\cite{sampat2009complex}, the gradient similarity (GSM) model~\cite{liu2011image}, the gradient magnitude similarity deviation (GMSD)~\cite{xue2013gradient}, and the superpixel-based similarity (SPSIM) index~\cite{sun2018spsim}. Other successful FR approaches include information theoretic models such as information fidelity criterion (IFC)~\cite{sheikh2005information} and visual information fidelity (VIF)~\cite{sheikh2006image}, and internal generative mechanism (IGM) model~\cite{wu2012perceptual}. NR methods may be derived from the natural scene statistics (NSS) models, and the notable ones include the blind/referenceless image spatial quality evaluator (BRISQUE)~\cite{mittal2012no}, the natural image quality evaluator (NIQE)~\cite{mittal2012making}, the blind image integrity notator using DCT statistics (BLIINDS-II)~\cite{saad2012blind}, and the distortion identification-based image verity and integrity evaluation (DIIVINE) index~\cite{moorthy2011blind}. NR methods may also be constructed based on local binary pattern statistics~\cite{wu2015highly}, or deep learning based approaches~\cite{ma2017dipiq,ma2017end,ying2020patches}.

\begin{figure*}[t]
	\centering
	\includegraphics[width=16.0cm]{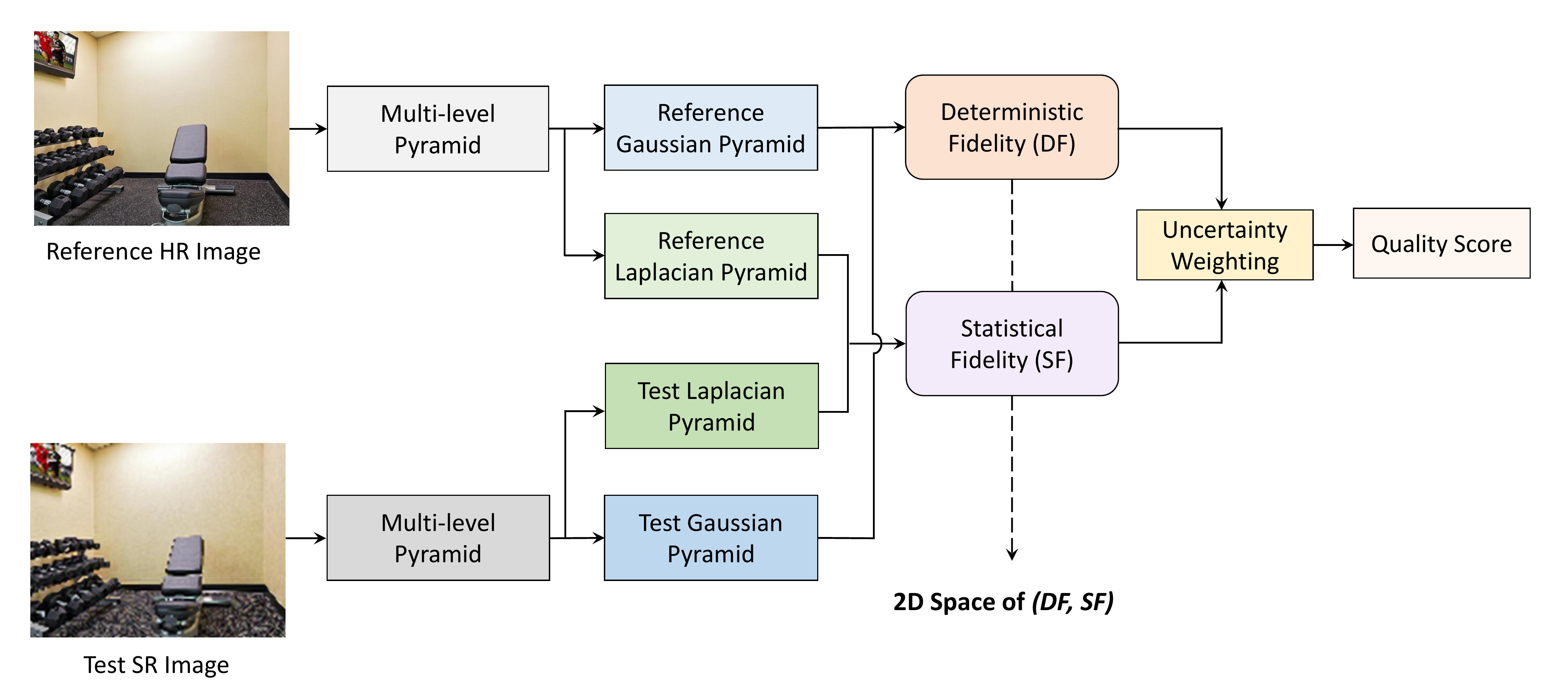}
	\caption{Framework of the proposed SRIF model.}
	\label{fig2}
\end{figure*}


Image sharpness/blurriness has been well studied in IQA research, which results in several successful methods including the spectral and spatial sharpness (S3)~\cite{vu2011bf} model, the local phase coherence-based sharpness index (LPC-SI) \cite{hassen2013image}, and the HVS-MaxPol models~\cite{hosseini2019encoding}. Nevertheless, they are limited in addressing only one perceptual attribute of SR images. There have been several approaches designed specifically for SR IQA, including the structure-texture decomposition based algorithm, i.e. SIS \cite{zhou2019visual}, and the structural fidelity versus statistical naturalness (SFSN) model \cite{zhou2021image}, though they do not directly differentiate the deterministic and statistical aspects of signal fidelity or employ content-adaptive weighting schemes.

\section{Proposed Method}
Assuming an original HR image is available as the reference, the framework of the proposed method is shown in Figure ~\ref{fig2}. Presumably, the perceptual deterministic and statistical fidelity varies across scales, and thus we first apply a multi-resolution decomposition to the reference and distorted images. 
Specifically, we apply a three-level Gaussian-Laplacian pyramid decomposition~\cite{burt1983laplacian} for simplicity. Empirically, we find that more sophisticated wavelet transforms also work well but do not offer additional gain.

Given a target image, a Gaussian pyramid is created by iterative Gaussian low-pass filtering and downsampling, resulting in a sequence of Gaussian maps denoted by ${G_l}(i,j)$ for $l=1,2,...$, where $l$ is the level of the Gaussian map. For each Gaussian map, an interpolation operation is then applied to expand the map by a factor of 2, leading to an expanded Gaussian map denoted by $G_l^*(i,j)$. A Laplacian map is then obtained by subtracting the expanded Gaussian map of the finer level from the Gaussian map of the current level by ${L_l} = {G_l} - G_{l + 1}^*$ for $l=1,2,...$, the collection of all such Laplacian maps constitute a Laplacian pyramid. The Gaussian pyramids of the reference and test images are used to evaluate the deterministic fidelity (DF), and the Laplacian pyramids of the reference and test images are employed to assess the statistical fidelity (SF). An uncertainty weighting scheme is then applied to produce a score that predicts the overall quality of the test image.

\subsection{Deterministic Fidelity}
To compute local DF, we resort to the structure comparison component of the SSIM approach \cite{wang2004image}, a simple but powerful tool to gauge the structural change between the reference and test image patches:
\begin{equation}
\centering
{D_{local}}(x,y) = \frac{{{\sigma _{xy}} + {C_1}}}{{{\sigma _x}{\sigma _y} + {C_1}}},
\end{equation}
where $x$ and $y$ are two local image patches extracted from the reference HR and test SR images, respectively, ${\sigma _x}$ and ${\sigma _y}$ denote their corresponding local standard deviations, ${\sigma _{xy}}$ represents the cross-correlation evaluation between the two patches, and $C_1$ is a small positive stabilizing constant. When applying this patch level measurement with a sliding window across one level of the Gaussian maps, which obtain a single-scale DF map.

\begin{figure*}[t]
	\centering
	\includegraphics[width=16cm]{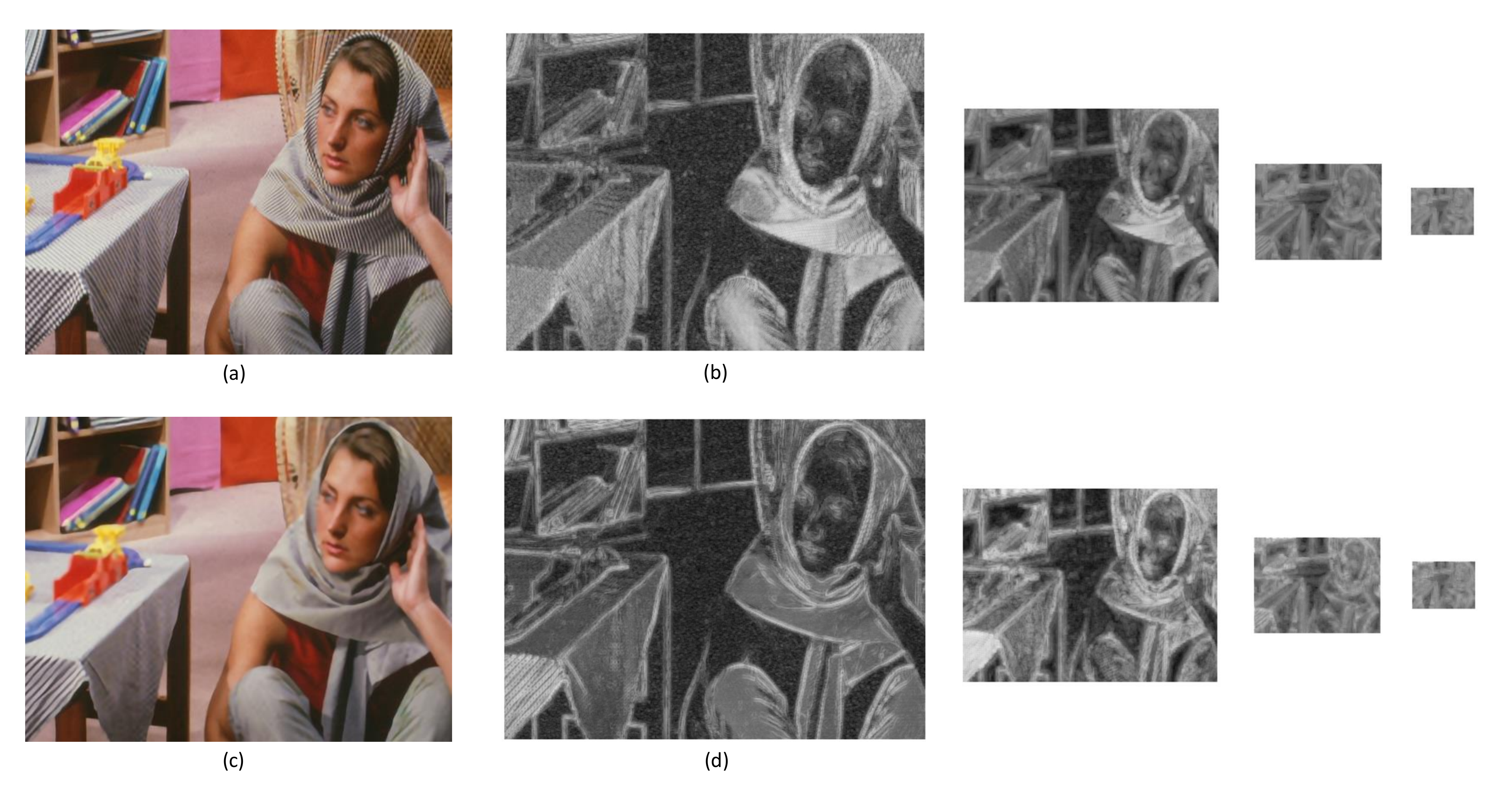}
	\caption{Sample information content maps. (a),(c): SR images generated by LapSRN \cite{lai2017deep} $\times$ 2 and $\times$ 4. (b),(d): Corresponding information content maps in a pyramid, where brighter indicates richer information content. The deterministic fidelity values of (a),(c) are 0.9862 and 0.9331, respectively.}
	\label{fig3}
\end{figure*}

Inspired by the success of the MS-SSIM and IW-SSIM methods~\cite{wang2003multiscale,wang2010information}, we adopt both scale weighting and information content pooling to aggregate the DF maps. For each level, we first employ an information content weighting to pool the single-scale DF map to a single DF score:
\begin{equation}
\centering
{D_{l,j}} = \frac{{\sum\limits_i {{w_{l,j,i}}{D_{local}}({x_{l,j,i}},{y_{l,j,i}})} }}{{\sum\limits_i {{w_{l,j,i}}} }},
\end{equation}
where ${x_{l,j,i}}$ and ${y_{l,j,i}}$ are the $i-th$ image patches in the $j-th$ scale for the $l-th$ level original and distorted Gaussian maps, respectively. ${w_{l,j,i}}$ is the information content weight~\cite{wang2010information} calculated at the $i-th$ spatial location. In Figure ~\ref{fig3}, we show an example of information content maps for two SR images. While the LapSRN~\cite{lai2017deep} $\times$ 2 method better preserves texture details compared to the same method $\times$ 4, its information content maps are brighter at sharp texture and edge regions. As such, stronger weights are assigned towards these content rich regions. Next, we combine all scale DF scores to a DF score for each level by
\begin{equation}
\centering
{D_l} = \prod\limits_{j = 1}^K {{{({D_{l,j}})}^{{\alpha _j}}}},
\end{equation}
where $K=5$ represents the total number of scales, and ${\alpha _j}$ denotes the weights assigned to the $j-th$ scale as in MS-SSIM~\cite{wang2003multiscale}. Finally, an overall DF measure is obtained by a weighted sum from all levels:
\begin{equation}
\centering
D = \sum\limits_l {{w_l}{D_l}},
\end{equation}
where $w_l$'s are weights assigned to individual levels.


\subsection{Statistical Fidelity}

let ${L_l}(i,j)$ denote the Laplacian map at the $l-th$ level, we normalize it locally by dividing it with the local standard deviation:
\begin{equation}
\centering
\widehat {{L_l}}(i,j) = \frac{{{L_l}(i,j) - \mu (i,j)}}{{\sigma (i,j) + C}},
\end{equation}
where $C$ is a small stabilizing constant, ${\mu _l}(i,j)$ and ${\sigma _l}(i,j)$ represent the mean and standard deviation of a local region surrounding ${{L_l}}(i,j)$, respectively, which in our experiment are computed using a local $3\times3$ window.



\begin{figure*}[t]
	\centering
	\includegraphics[width=16.0cm]{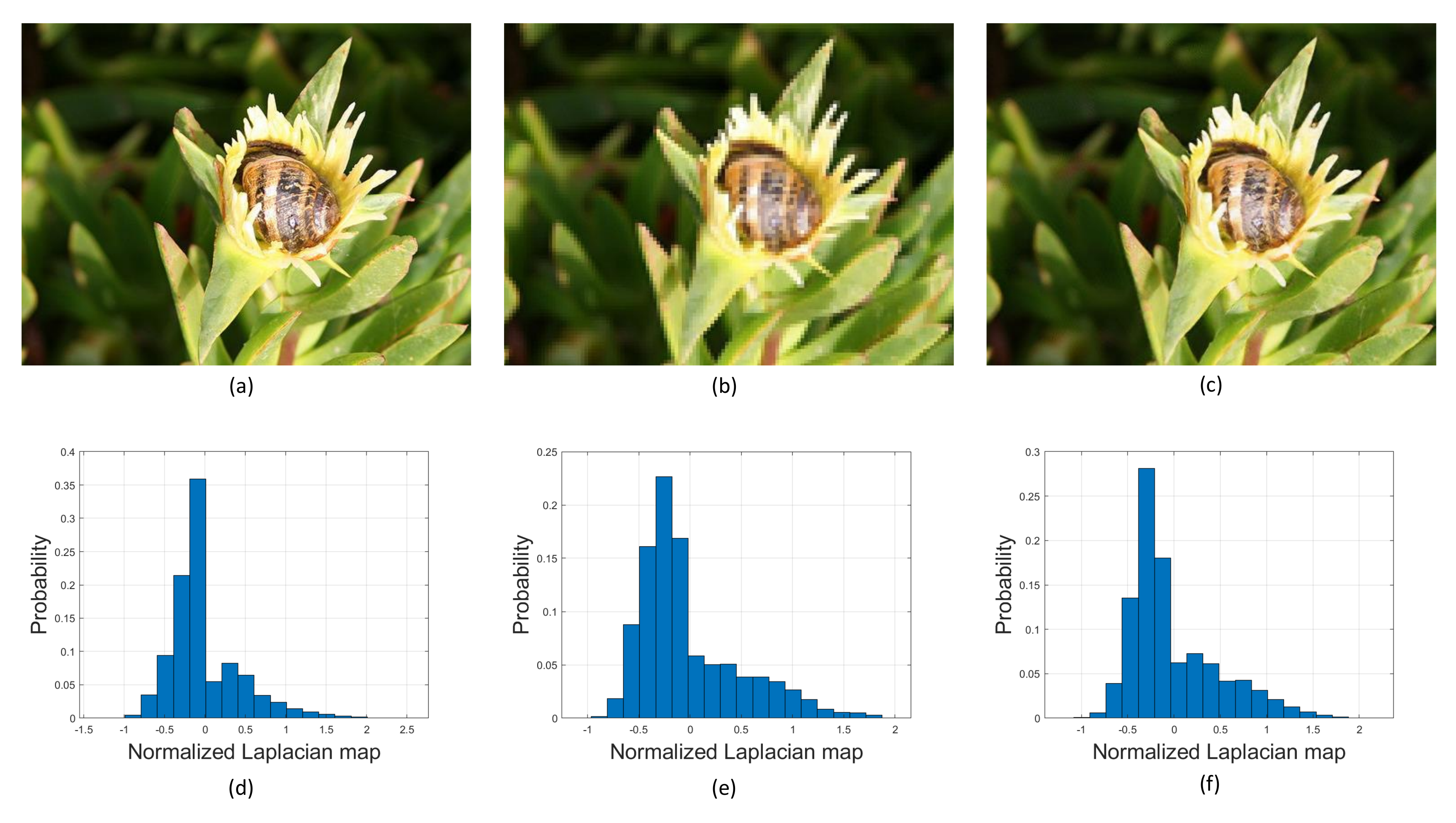}
	\caption{Illustration of SF assessment. (a-c): Reference and SR images produced by Example-based SR \cite{freeman2002example} and SRGAN \cite{ledig2017photo}, respectively; (d-f): Corresponding histograms of normalized Laplacian pyramid. The KLDs between (d) and (e) and between (d) and (f) are 0.0162 and 0.0138, respectively, suggesting SR image (c) better preserves SF.}
	\label{fig4}
\end{figure*}

Now assume that $p_l(x)$ and $q_l(x)$ are the probability densities of the normalized reference and test Laplacian maps at the $l-th$ level, respectively. We then calculate the Kullback-Leibler divergence (KLD) between $p_l(x)$ and $q_l(x)$ to measure their difference:
\begin{equation}
\centering
{S_l}(p||q) = \int {{p_l}(x)\log \frac{{{p_l}(x)}}{{{q_l}(x)}}dx}.
\end{equation}
An example is given in Figure ~\ref{fig4}, where we show the histograms of normalized Laplacian maps of the original HR image and two test SR images produced by two SR algorithms. Perceptually the SRGAN~\cite{ledig2017photo} generated SR image reveals more texture details as in the reference image, outperforming the example-based SR method~\cite{freeman2002example}. Meanwhile, the histogram of the SRGAN image appears more similar to that of the original image, which is well reflected by the KLD measure, suggesting KLD is a promising indicator of SF.
Finally, the overall SF measure is computed by a weighted sum of SF from each level:
\begin{equation}
\centering
S = \sum\limits_l {{w_l}{S_l}},
\end{equation}
where $w_l$'s are weights assigned to individual levels.


\subsection{Uncertainty Weighting}
Both DF and SF are sensible measures that when working individually provides meaningful assessment about the intrinsic characteristics of SR algorithms under testing. In certain application scenarios, however, one desires to have a single quality or fidelity score that considers both fidelity measures. A direct combination, e.g., a weighted average with fixed weights, would not be an optimal solution because the importance and uncertainty of the DF and SF measures may be content dependent. For example, different content exhibits different sharpness and texture richness, which lead to different levels of uncertainty when DF and SF are employed as the overall quality predictors.



We extract sharpness and texture richness features using the same Gaussian and Laplacian pyramids created for the reference and test images. The LPC-SI approach~\cite{hassen2013image} is first employed for sharpness assessment:
\begin{equation}
\centering
sr = \frac{{{l_{d - G1}}}}{{{l_{o - G2}}}},
\end{equation}
where ${l_{d - G1}}$ and ${l_{o - G2}}$ denote the LPC-SI values computed from the first level of the Gaussian pyramid of the test image and the second level of the Gaussian pyramid of the original image, respectively. We then use entropy as a measure for texture richness and define a texture richness ratio as
\begin{equation}
\centering
tr = \frac{{{e_{d - L1}}}}{{{e_{o - G2}}}},
\end{equation}
where ${e_{d - L1}}$ and ${e_{o - G2}}$ represent the entropy of the first level of the normalized Laplacian map of the test image and that of the second level of the reference Gaussian pyramid, respectively. The sharpness and texture richness ratios are then integrated to an assorted factor as
\begin{equation}
\centering
f = {(sr)^\alpha } + {(tr)^\alpha },
\end{equation}
where an exponential factor $\alpha$ is added to stretch the value range.

\begin{figure}[t]
	\centering
	\includegraphics[width=8.8cm]{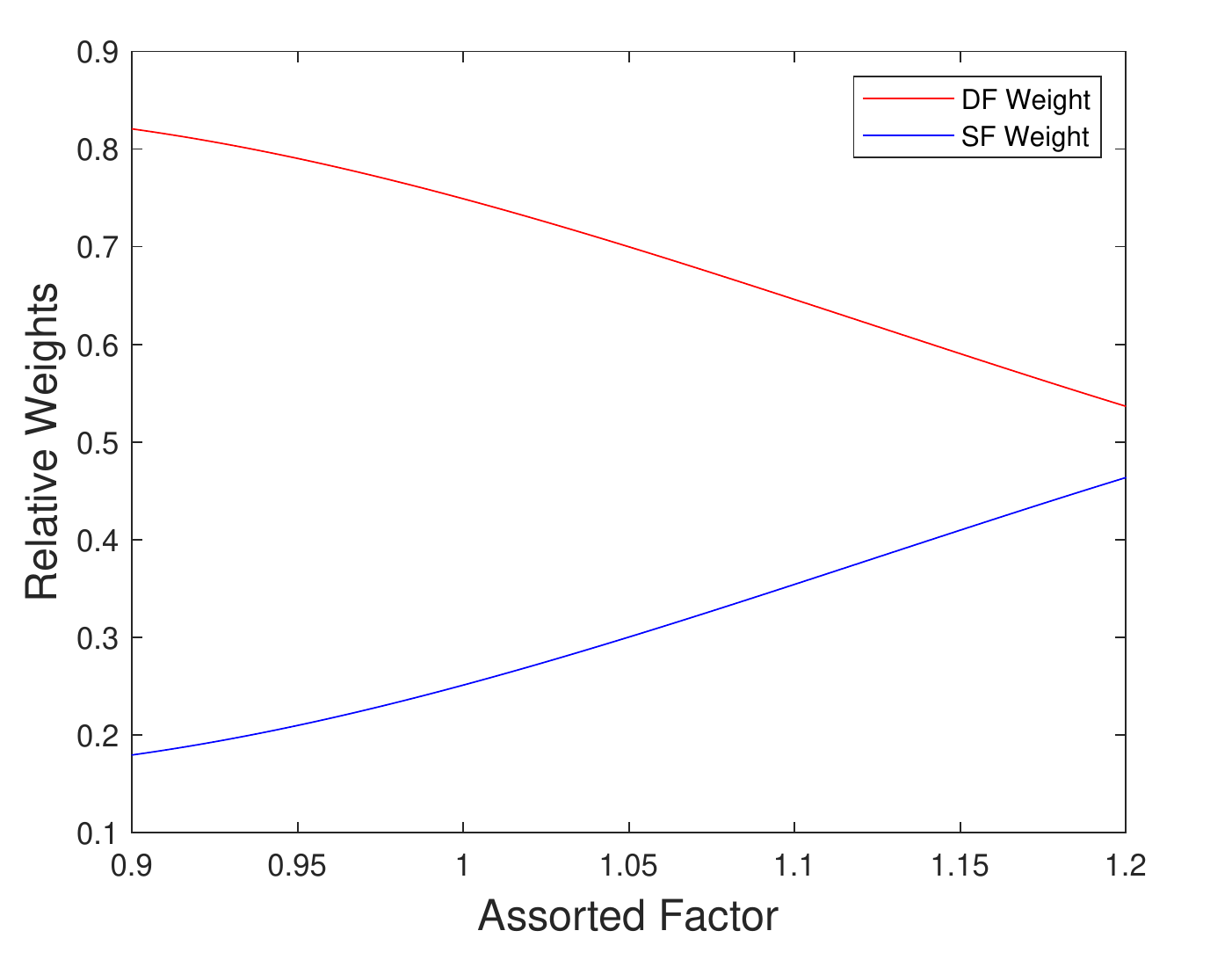}
	\caption{Weights assigned to DF and SF measures with respect to the assorted factor.}
	\label{fig5}
\end{figure}

\begin{figure}[t]
	\centering
	\includegraphics[width=8.8cm]{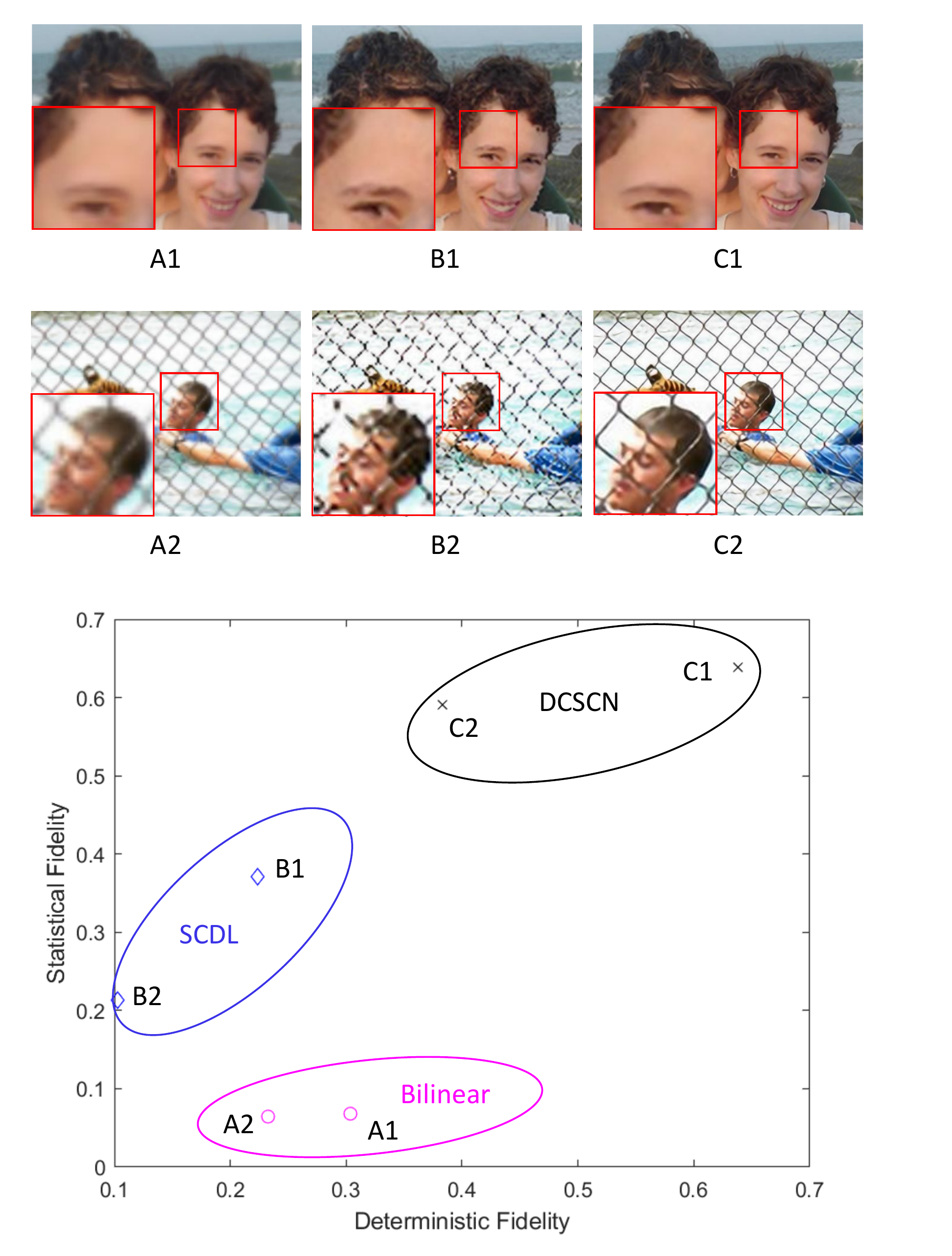}
\caption{Analysis of different SR algorithms in (DF, SF) space. A1-A2, B1-B2, and C1-C2: SR images created by Bilinear interpolation, SCDL \cite{wang2012semi}, and DCSCN \cite{yamanaka2017fast} at scaling factor 4, respectively. Bottom: Clusters of SR algorithms in (DF, SF) space.}
	\label{fig6}
\end{figure}

\begin{figure}[t]
	\centering
	\includegraphics[width=8.8cm]{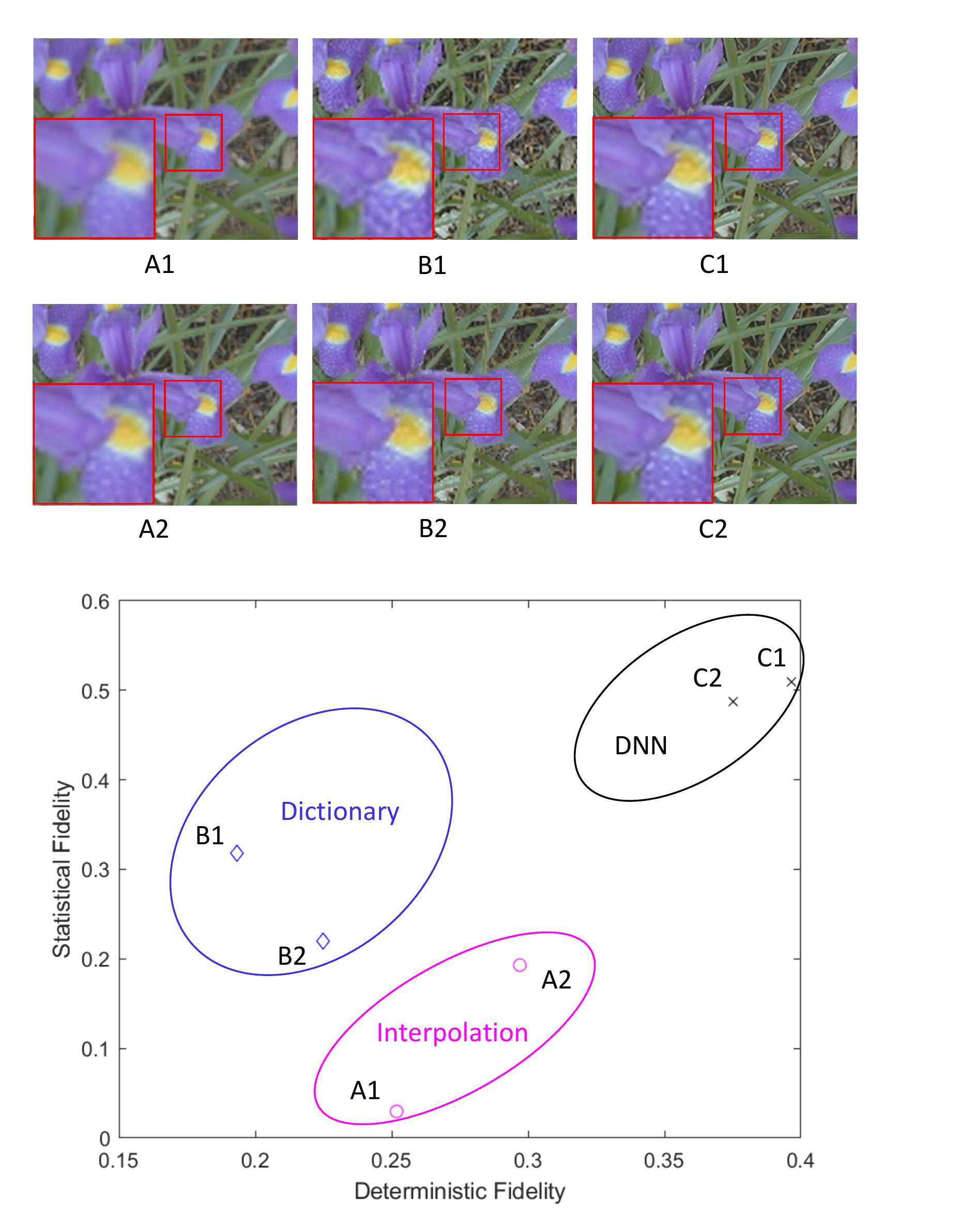}
	\caption{Analysis of various SR categories in (DF, SF) space. A1-A2: SR images reconstructed by Bilinear and Bicubic interpolations; B1-B2: SR images produced by SCDL \cite{wang2012semi} and Fast SR \cite{yang2013fast}; C1-C2: SR images generated by DCSCN \cite{yamanaka2017fast} and LapSRN \cite{lai2017deep}. The scaling factor is 4 for all cases. Bottom: Clusters of SR categories in (DF, SF) space.}
	\label{fig7}
\end{figure}

To evaluate how the prediction uncertainty of the DF and SF measures changes with respect to the assorted factor, we bin images along the axis of the assorted factor and compute the variances of the prediction errors for DF and SF within each bin. Presumably, a larger variance implies larger uncertainty, and thus should carry less weight in the overall prediction. Therefore, we compute the uncertainty weighting factors for DF and SF at each bin by
\begin{equation}
\centering
{w_d} = \frac{{{v_s}}}{{{v_d} + {v_s}}},
\end{equation}
\begin{equation}
\centering
{w_s} = \frac{{{v_d}}}{{{v_d} + {v_s}}},
\end{equation}
where $v_d$ and $v_s$ are the variances of the prediction errors of the DF and SF measures, respectively. Figure~\ref{fig5} shows how these weighting factors change against the assorted factor. It can be observed that generally DF carries more weights than SF, but as the assorted factor increases, more weights gradually shift towards the SF measure, which is intuitively sensible because an image with larger assorted factor contains richer textures that are better accounted for by statistical measurements. Finally, the overall uncertainty-weighted Super Resolution Image Fidelity (SRIF) index is calculated by
\begin{equation}
\centering
{Q_{ds}} = {w_d}D + {w_s}S\,.
\end{equation}



\section{Validation}
\label{sec:validation}

We validate the proposed method and compare it with other objective quality models using three publicly available SR quality datasets, including the Waterloo Interpolation~\cite{yeganeh2015objective}, QADS~\cite{zhou2019visual} and CVIU~\cite{ma2017learning} databases. The Waterloo Interpolation~\cite{yeganeh2015objective} database involves 8 interpolation algorithms with 3 interpolation factors of 2, 4, and 8, respectively. 312 SR images are generated from 13 source images. The QADS~ \cite{zhou2019visual} database contains 20 original HR images and 980 SR images created by 21 image SR algorithms, including 4 interpolation-based, 11 dictionary-based, and 6 DNN-based SR models, with upsampling factors equaling to 2, 3, and 4. 
Each SR image is associated with a mean opinion score (MOS) of 100 subjects. 
In the CVIU database~\cite{ma2017learning}, 1620 SR images are produced by 9 SR approaches from 30 source images. 
6 pairs of scaling factors and kernel widths are adopted, where a larger subsampling factor corresponds to a larger blur kernel width. Each image is rated by 50 subjects, and the mean of the median 40 scores is calculated for each image as the MOS.

\begin{figure}[t]
	\centering
	\includegraphics[width=8.0cm]{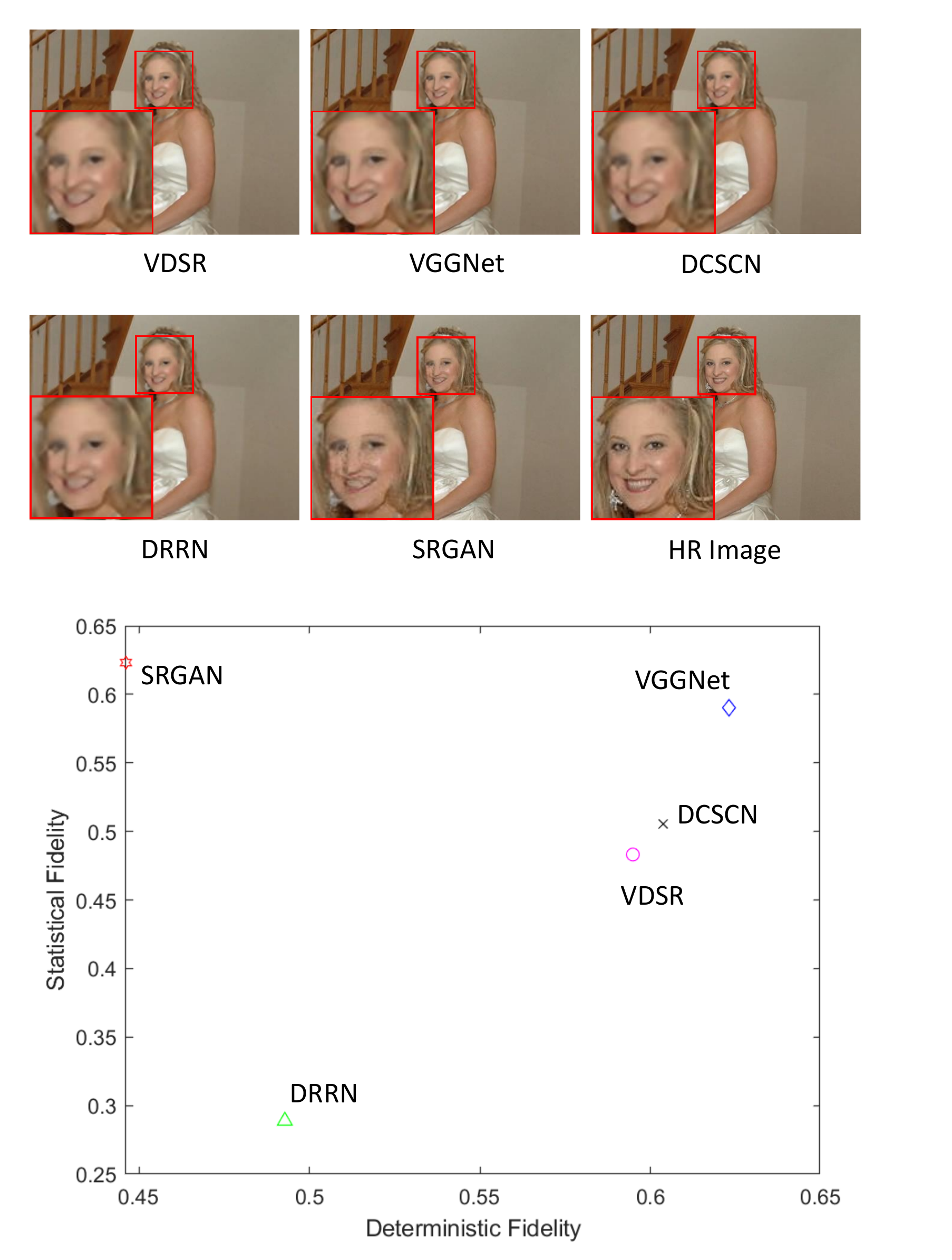}
	\caption{Analysis of DNN-based SR algorithms in (DF, SF) space. VDSR \cite{kim2016accurate}, VGGNet \cite{simonyan2014very}, DCSCN \cite{yamanaka2017fast} and DRRN \cite{tai2017image} are CNN-based methods, while SRGAN \cite{ledig2017photo} is GAN-based method. Bottom: Quality assessment in (DF, SF) space.}
	\label{fig8}
\end{figure}

\subsection{2D Analysis of SR Algorithms}
As mentioned earlier, the behaviors of different SR algorithms may be drastically different, and it is difficult to use one scalar number to describe the visual quality of images generated by these algorithms. The proposed 2D (DF, SF) framework provides a useful tool to analyze the characteristics of the SR algorithms.


Figure~\ref{fig6} shows such an example in comparing different SR algorithms and analyzing their strengths and weaknesses. It can be observed that Bilinear Interpolation blurs the image, loses the texture details, and severely deteriorates the statistical properties or SF of the original image. Even though the SCDL \cite{wang2012semi} generated image appears a bit better on the statistical properties, it performs poorly in preserving the authentic structures or DF of the original image. By contrast, the DCSCN~\cite{yamanaka2017fast} algorithm offers impressive performance in preserving both deterministic structures and statistical properties of the original image. All of these observations are clearly indicated in the 2D (DF, SF) space.



It is also interesting to observe the behaviors of different categories of SR algorithms. In particular, there are totally 21 SR algorithms in the QADS database~\cite{zhou2019visual}, which are divided into three categories: 4 interpolation-based, 11 dictionary-based, and 6 DNN-based methods. In Figure ~\ref{fig7}, we carry out a similar performance analysis and comparison in the (DF, SF) space but on different categories of SR algorithms. The results reveal complementary advantages of interpolation and dictionary-based SR algorithms, and demonstrates the strong performance of DNN-based methods. The conclusion aligns very well with visual observations of the reconstructed SR images. We further analyze various DNN-based SR algorithms, as shown in Figure~\ref{fig8}. The general trends of CNN-based SR models are consistent, where DF and SF are inclined to improve together. But the behavior of GAN-based SR method (i.e. SRGAN) is dramatically different, which tends to sacrifice DF for SF.



\subsection{Performance Evaluation and Comparison}

Given a dataset containing SR images and their corresponding subjective quality scores, we evaluate and compare objective models for their quality prediction performance using four criteria, namely Spearman Rank-order Correlation Coefficient (SRCC), Kendall Rank-order Correlation Coefficient (KRCC), Pearson Linear Correlation Coefficient (PLCC), and Root Mean Squared Error (RMSE). SRCC and KRCC are used to assess prediction monotonicity, and PLCC and RMSE are employed to evaluate prediction accuracy. Before PLCC and RMSE are computed, a five-parameter modified logistic nonlinear fitting function \cite{rohaly2000video} is employed to map the predicted quality scores into a linear scale of the subjective scores by
\begin{equation}
g(x) = {\beta _1}(\frac{1}{2} - \frac{1}{{1 + {e^{({\beta _2}(x - {\beta _3}))}}}}) + {\beta _4}x + {\beta _5},
\end{equation}
where $({\beta _1}$ to ${\beta _5})$ are fitting parameters, and $x$ and $g(x)$ represent the raw and regressed objective scores, respectively.

\renewcommand\arraystretch{1.0}
\begin{table*}[t]
	\centering
	\scriptsize
	\caption{Performance comparison on Waterloo Interpolation \cite{yeganeh2015objective}, QADS \cite{zhou2019visual} and CVIU \cite{ma2017learning} quality databases.}
	\begin{tabular}{c|c|cccc|cccc|cccc}
		\hline
		\multicolumn{1}{c|}{} & \multicolumn{1}{c|}{} & \multicolumn{4}{c|}{\textbf{Waterloo Interpolation}} & \multicolumn{4}{c|}{\textbf{QADS}} & \multicolumn{4}{c}{\textbf{CVIU}}\\ \hline
		\textbf{Types} & \textbf{Methods} & \textbf{SRCC} & \textbf{KRCC} & \textbf{PLCC} & \textbf{RMSE} & \textbf{SRCC} & \textbf{KRCC} & \textbf{PLCC} & \textbf{RMSE} & \textbf{SRCC} & \textbf{KRCC} & \textbf{PLCC} & \textbf{RMSE} \\ \hline
		\multirow{11}{*}{\textbf{FR IQA Models}} & \textbf{PSNR} &0.6320 &0.4421 &0.6303 &2.0020 &0.3544 &0.2441 &0.3897 &0.2530 &0.5663 &0.3943 &0.5779 &1.9620 \\
		& \textbf{SSIM \cite{wang2004image}} &0.6125 &0.4311 &0.6208 &2.0215 &0.5290 &0.3689 &0.5327 &0.2325 &0.6285 &0.4429 &0.6497 &1.8277 \\
		& \textbf{MS-SSIM \cite{wang2003multiscale}} &0.8246 &0.6228 &0.8371 &1.4108 &0.7172 &0.5299 &0.7240 &0.1895 &0.8048 &0.6011 &0.8114 &1.4052 \\
		& \textbf{FSIM \cite{zhang2011fsim}} &0.8503 &0.6474 &0.8595 &1.3179 &0.6885 &0.5020 &0.6902 &0.1988 &0.7481 &0.5450 &0.7628 &1.5546 \\
		& \textbf{CW-SSIM \cite{sampat2009complex}} &0.8626 &0.6658 &0.9056 &1.0935 &0.3259 &0.2275 &0.3792 &0.2542 &0.7591 &0.5410 &0.7541 &1.5790 \\
        & \textbf{IFC \cite{sheikh2005information}} &0.9117	&0.7227	&0.9350	&0.9146 &0.8609	&0.6816	&0.8657	&0.1375	&0.8705	&0.6807	&0.8836	&1.1259 \\
        & \textbf{VIF \cite{sheikh2006image}} &0.8928 &0.6973 &0.9053 &1.0953 &0.8152	&0.6249	&0.8210	&0.1568	&0.8571	&0.6630	&0.8671	&1.1974 \\
        & \textbf{GSM \cite{liu2011image}} &0.7649 &0.5650 &0.7773 &1.6224 &0.5538	&0.3908	&0.5684	&0.2260	&0.6505	&0.4587	&0.6717	&1.7809 \\
        & \textbf{IGM \cite{wu2012perceptual}} &0.8659 &0.6675 &0.8841 &1.2051 &0.7145	&0.5232	&0.7192	&0.1909	&0.8375	&0.6375	&0.8442	&1.2885 \\
        & \textbf{GMSD \cite{xue2013gradient}} &0.7966 &0.5915 &0.8108 &1.5092 &0.7650	&0.5689	&0.7749	&0.1736	&0.8469	&0.6495	&0.8499	&1.2669 \\
		& \textbf{SPSIM \cite{sun2018spsim}} &0.8141 &0.6143 &0.8263 &1.4524 &0.5751 &0.4071 &0.5829 &0.2232 &0.6698 &0.4755 &0.6902 &1.7396 \\ \hline
		\multirow{8}{*}{\textbf{NR IQA Models}} & \textbf{BRISQUE \cite{mittal2012no}} &0.7676	&0.5610	&0.7871	&1.5907 &0.5463	&0.3834	&0.5478	&0.2298	&0.5863	&0.4262	&0.6094	&1.9061 \\
        & \textbf{NIQE \cite{mittal2012making}} &0.6263	&0.4647	&0.6716	&1.9105 &0.3977	&0.2788	&0.4044	&0.2512	&0.6525	&0.4775	&0.6657	&1.7940 \\
        & \textbf{BLIINDS-II \cite{saad2012blind}} &0.5281 &0.3531 &0.5260 &2.1932 &0.3838 &0.2617 &0.4490 &0.2454 &0.3705 &0.2584 &0.4389	&2.1602 \\
        & \textbf{DIIVINE \cite{moorthy2011blind}} &0.5465 &0.3778 &0.5719 &2.1158 &0.4817	&0.3314	&0.5044	&0.2372	&0.5479	&0.3956	&0.6347 &1.8577 \\
        & \textbf{LPSI \cite{wu2015highly}} &0.6669	&0.4642	&0.7007	&1.8396 &0.4079	&0.2890	&0.4217	&0.2491	&0.4883	&0.3503	&0.5374	&2.0274 \\
        & \textbf{dipIQ \cite{ma2017dipiq}} &0.6560	&0.4617	&0.6727	&1.9080 &0.5057	&0.3529	&0.5296	&0.2330	&0.4901	&0.3640	&0.5212	&2.0518 \\
		& \textbf{MEON \cite{ma2017end}} &0.8554 &0.6286 &0.8672 &1.2843 &0.6139 &0.4528 &0.6162 &0.2163 &0.4599 &0.3253 &0.5124 &2.0645  \\
		& \textbf{PaQ-2-PiQ \cite{ying2020patches}} &0.7479 &0.5214	&0.7808	&1.6112	&0.7473	&0.5465	&0.7514	&0.1812	&0.7294	&0.5403	&0.7376	&1.6233\\ \hline
		\multirow{4}{*}{\textbf{Sharpness Models}} & \textbf{S3 \cite{vu2011bf}} &0.4455 &0.3099 &0.4999 &2.2334 &0.4636 &0.3198 &0.4720 &0.2422 &0.5050 &0.3519 &0.5480 &2.0109 \\
		& \textbf{LPC-SI \cite{hassen2013image}} &0.5375 &0.3725 &0.5944 &2.0736 &0.4902 &0.3358 &0.5027 &0.2374 &0.5450 &0.3921 &0.5731 &1.9701 \\
		& \textbf{HVS-MaxPol-1 \cite{hosseini2019encoding}} &0.6166	&0.4329	&0.6577	&1.9425 &0.6170	&0.4322	&0.6212	&0.2153	&0.6421	&0.4661	&0.6703	&1.7840 \\
		& \textbf{HVS-MaxPol-2 \cite{hosseini2019encoding}} &0.6309	&0.4284	&0.6680	&1.9190 &0.5736	&0.3986	&0.5847	&0.2228	&0.6313	&0.4517	&0.6474	&1.8324 \\ \hline
		\multirow{3}{*}{\textbf{SR IQA Models}} &\textbf{SIS \cite{zhou2019visual}} &0.8777 &0.6773 &0.8913 &1.1692 &0.9132 &0.7397 &0.9137 &0.1116 &0.8694 &0.6855 &0.8973 &1.0611 \\
		&\textbf{SFSN \cite{zhou2021image}} &0.8867 &0.6917 &0.9058 &1.0928 &0.8407 &0.6553 &0.8445 &0.1471 &0.8714 &0.6800 &0.8848 &1.1202 \\
		& \textbf{SRIF (proposed)} &\textbf{0.9157} &\textbf{0.7299} &\textbf{0.9525} &\textbf{0.7855} &\textbf{0.9163} &\textbf{0.7457} &\textbf{0.9174} &\textbf{0.1093}	&\textbf{0.8857} &\textbf{0.7042} &\textbf{0.9018} &\textbf{1.0389} \\ \hline
	\end{tabular}
\label{table1}
\end{table*}

In addition to SRIF, a total of 25 other models are included in performance comparison. These include eleven FR IQA, eight NR IQA, four sharpness models, and two SR IQA approaches.

The results using aforementioned datasets and evaluation criteria are reported in Table~\ref{table1}, where the top performers of the FR IQA, NR IQA and sharpness model categories are IFC~\cite{sheikh2005information}, MEON~\cite{ma2017end} or PaQ-2-PiQ~\cite{ying2020patches}, and HVS-MaxPol~\cite{hosseini2019encoding}, respectively, but in general, they underperform the models dedicated to SR IQA. Of all the models under comparison, the proposed SRIF model produces the best performance for all evaluation criteria in all three databases. 


To investigate deeper on the performance of objective quality models for different categories of SR algorithms, we report SRCC results of these models separately for interpolation-based, dictionary-based and DNN-based SR approaches on the QADS dataset~\cite{zhou2019visual} in Table~\ref{table2}. It is interesting to observe that FR IQA methods usually performs better in evaluating dictionary-based SR algorithms, while most NR IQA models show relatively better performance on DNN-based against the other two categories of SR approaches. The proposed SRIF model offers consistently good performance across all three categories.

\renewcommand\arraystretch{1.0}
\begin{table*}[t]
	\centering
	\scriptsize
	\caption{SRCC Performance comparison of objective quality models on QADS \cite{zhou2019visual} database.}
	\begin{tabular}{c|c|cccc}
		\hline
		\textbf{Types} & \textbf{Methods} & \textbf{Interpolation-based} & \textbf{Dictionary-based} & \textbf{DNN-based} & \textbf{All} \\ \hline
		\multirow{11}{*}{\textbf{FR IQA Models}} & \textbf{PSNR}           &0.2972	&0.3808	&0.2656	&0.3544 \\
		& \textbf{SSIM \cite{wang2004image}}                                             &0.4015	&0.5481	&0.5121	&0.5290 \\
		& \textbf{MS-SSIM \cite{wang2003multiscale}}                                          &0.6340	&0.7425	&0.7104	&0.7172 \\
		& \textbf{FSIM \cite{zhang2011fsim}}                                             &0.5471	&0.6846	&0.6637	&0.6885 \\
		& \textbf{CW-SSIM \cite{sampat2009complex}}                                          &0.5254	&0.4362	&0.0986	&0.3259 \\
        & \textbf{IFC \cite{sheikh2005information}}                                              &0.8489	&0.8835	&0.7792	&0.8609 \\
        & \textbf{VIF \cite{sheikh2006image}}                                              &0.7654	&0.8305	&0.7281	&0.8152 \\
        & \textbf{GSM \cite{liu2011image}}                                              &0.3946	&0.5332	&0.5661	&0.5538 \\
        & \textbf{IGM \cite{wu2012perceptual}}                                              &0.6285	&0.7292	&0.6625	&0.7145 \\
        & \textbf{GMSD \cite{xue2013gradient}}                                             &0.7054	&0.7709	&0.7363	&0.7650 \\
		& \textbf{SPSIM \cite{sun2018spsim}}                                            &0.4545	&0.5518	&0.5871	&0.5751 \\ \hline
		\multirow{8}{*}{\textbf{NR IQA Models}} & \textbf{BRISQUE \cite{mittal2012no}} &0.5096	&0.4951	&0.4357	&0.5463 \\
        & \textbf{NIQE \cite{mittal2012making}}                                             &0.4639	&0.4547	&0.4190	&0.3977 \\
        & \textbf{BLIINDS-II \cite{saad2012blind}}                                       &0.1814	&0.3628	&0.6547	&0.3838 \\
        & \textbf{DIIVINE \cite{moorthy2011blind}}                                          &0.4267	&0.4175	&0.5654	&0.4817 \\
        & \textbf{LPSI \cite{wu2015highly}}                                             &0.2726	&0.3309	&0.6034	&0.4079 \\
        & \textbf{dipIQ \cite{ma2017dipiq}}                                            &0.3368	&0.4980	&0.4021	&0.5057 \\
		& \textbf{MEON \cite{ma2017end}}                                             &0.4809	&0.5956	&0.6951	&0.6139  \\
		& \textbf{PaQ-2-PiQ \cite{ying2020patches}} &0.7272 &0.6853 &0.5533 &0.7473 \\ \hline
		\multirow{4}{*}{\textbf{Sharpness Models}} & \textbf{S3 \cite{vu2011bf}}        &0.4016	&0.3171	&0.5458	&0.4636 \\
		& \textbf{LPC-SI \cite{hassen2013image}}                            &0.3301	&0.3798	&0.2558	&0.4902 \\
		& \textbf{HVS-MaxPol-1 \cite{hosseini2019encoding}}                      &0.4584	&0.5048	&0.5032	&0.6170 \\
		& \textbf{HVS-MaxPol-2 \cite{hosseini2019encoding}}                &0.5318	&0.4742	&0.2991	&0.5736 \\ \hline
		\multirow{3}{*}{\textbf{SR IQA Models}} &\textbf{SIS \cite{zhou2019visual}} &0.8778	&\textbf{0.9038} &0.8493	&0.9132 \\
		&\textbf{SFSN \cite{zhou2021image}} &0.8979	&0.8379 &0.8004	&0.8407 \\
		& \textbf{SRIF (proposed)}                                            &\textbf{0.9119} &0.8950 &\textbf{0.8778} &\textbf{0.9163} \\ \hline
	\end{tabular}
\label{table2}
\end{table*}

\renewcommand\arraystretch{1.0}
\begin{table*}[t]
	\centering
    \scriptsize
	\caption{Ablation test of the proposed SRIF model on Waterloo Interpolation \cite{yeganeh2015objective}, QADS \cite{zhou2019visual} and CVIU \cite{ma2017learning} databases.}
	\begin{tabular}{c|cccc|cccc|cccc}
		\hline
	& \multicolumn{4}{c|}{\textbf{Waterloo Interpolation}}	& \multicolumn{4}{c|}{\textbf{QADS}}       & \multicolumn{4}{c}{\textbf{CVIU}}   \\ \hline
		\textbf{Models} & \textbf{SRCC} & \textbf{KRCC} & \textbf{PLCC} & \textbf{RMSE} & \textbf{SRCC} & \textbf{KRCC} & \textbf{PLCC} & \textbf{RMSE} & \textbf{SRCC} & \textbf{KRCC} & \textbf{PLCC} & \textbf{RMSE} \\ \hline
		\textbf{DF only}   & 0.9152 & 0.7265 & 0.9517 & 0.7918 &0.7813 &0.5929 &0.7890 &0.1688 &0.8529 &0.6547 &0.8702 &1.1844 \\
		\textbf{SF only}   & 0.8147 & 0.5867 & 0.8414 & 1.3936 &0.8480 &0.6625 &0.8501 &0.1446 &0.8154 &0.6169 &0.8246 &1.3601 \\
        \textbf{SRIF} &\textbf{0.9157} &\textbf{0.7299} &\textbf{0.9525} &\textbf{0.7855} &\textbf{0.9163}	&\textbf{0.7457} &\textbf{0.9174} &\textbf{0.1093} &\textbf{0.8857} &\textbf{0.7042} &\textbf{0.9018} &\textbf{1.0389} \\ \hline
	\end{tabular}
	\label{table3}
\end{table*}

\renewcommand\arraystretch{1.0}
\begin{table*}[t]
	\centering
    \scriptsize
	\caption{Comparison of weighting strategies on Waterloo Interpolation \cite{yeganeh2015objective}, QADS \cite{zhou2019visual} and CVIU \cite{ma2017learning} databases.}
	\begin{tabular}{c|cccc|cccc|cccc}
		\hline
		\multicolumn{1}{c|}{} & \multicolumn{4}{c|}{\textbf{Waterloo Interpolation}} & \multicolumn{4}{c|}{\textbf{QADS}}  & \multicolumn{4}{c}{\textbf{CVIU}} \\ \hline
		\textbf{Methods} & \textbf{SRCC} & \textbf{KRCC} & \textbf{PLCC} & \textbf{RMSE} & \textbf{SRCC} & \textbf{KRCC} & \textbf{PLCC} & \textbf{RMSE} & \textbf{SRCC} & \textbf{KRCC} & \textbf{PLCC} & \textbf{RMSE} \\ \hline
		\textbf{Average Weighting} &0.8908 &0.6860 &0.9287 &0.9564 &0.8863 &0.7092 &0.8906 &0.1249 &0.8750 &0.6857 &0.8870 &1.1101 \\
        \textbf{Proposed Uncertainty Weighting} &\textbf{0.9157} &\textbf{0.7299} &\textbf{0.9525} &\textbf{0.7855} &\textbf{0.9163}	&\textbf{0.7457} &\textbf{0.9174} &\textbf{0.1093} &\textbf{0.8857} &\textbf{0.7042} &\textbf{0.9018} &\textbf{1.0389} \\ \hline
	\end{tabular}
	\label{table4}
\end{table*}

\subsection{Ablation Tests}
It is desirable to know how the individual DF and SF components attribute to the overall performance of the SRIF model. In Table~\ref{table3}, we compare the performance of the DF only model, the SF only model, and the overall SRIF model on the Waterloo Interpolation \cite{yeganeh2015objective}, QADS~\cite{zhou2019visual} and CVIU~\cite{ma2017learning} databases. It can be seen that for the Waterloo Interpolation dataset, where only interpolation is considered, DF can deliver relatively good performance. This is reasonable since interpolation-based SR algorithms often blur the image, which protects the primary structure. However, when a variety of SR categories are involved, such as the QADS and CVIU datasets, the DF only model is insufficient. In general, we show that both the DF and SF components offer competitive performance individually, but the overall SRIF model elevates the performance to an even higher level, suggesting that the DF and SF components, to some extent, complement each other.



\begin{figure}[t]
	\centering
	\includegraphics[width=9.0cm]{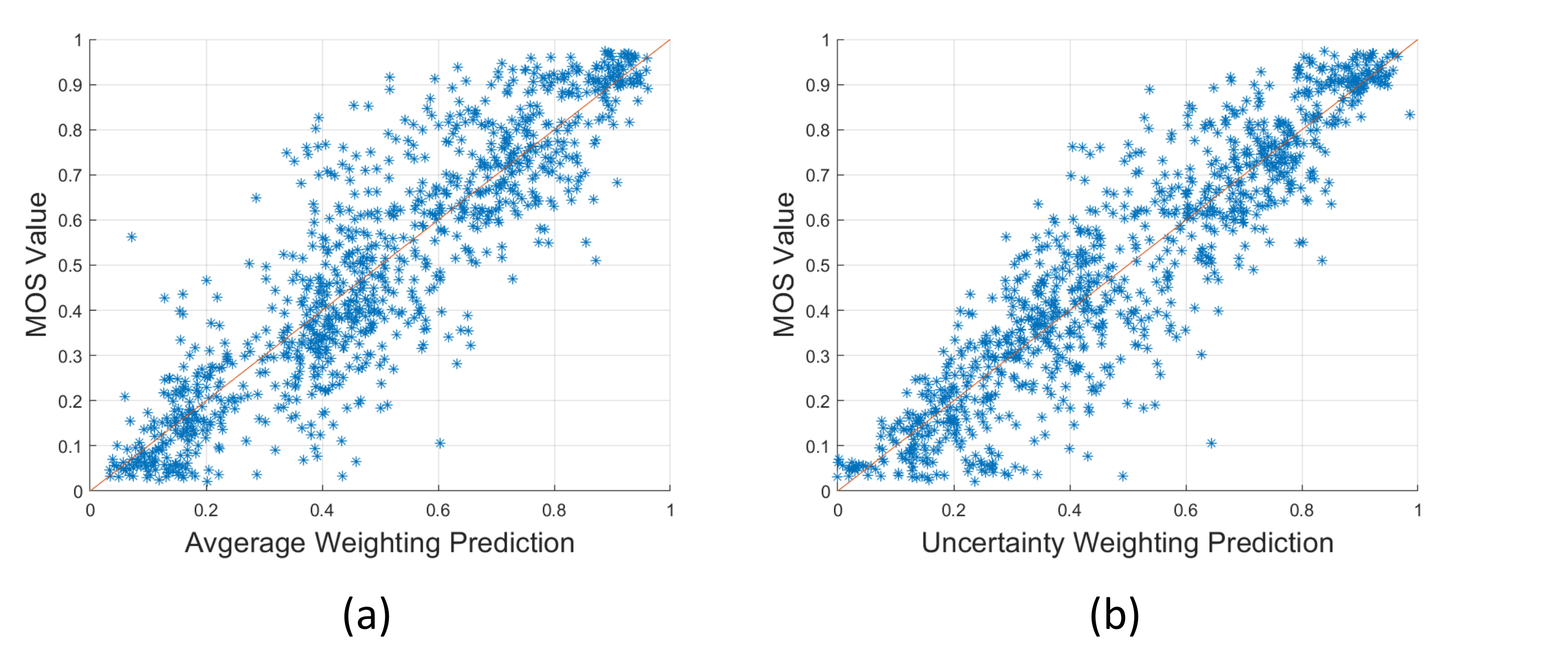}
	\caption{Scatter plots of subjective scores against objective model predictions with different weighting strategies. (a) average weighting; (b) proposed uncertainty weighting.}
	\label{fig9}
\end{figure}

To understand the contribution of the uncertainty weighting scheme in SRIF, we compare SRIF with a na\"ive model that directly averages the DF and SF components. Table~\ref{table4} shows the comparison results based on the Waterloo Interpolation~\cite{yeganeh2015objective}, QADS~\cite{zhou2019visual} and CVIU~\cite{ma2017learning} databases. The proposed uncertainty weighting scheme performs better in all test cases. Moreover, we show the scatter plots in Figure~\ref{fig9} of MOS versus model predictions using average and uncertainty weighting schemes, respectively, where each point represents a test SR image. It appears that the points in the scatter plot corresponding to uncertainty weighting are more tightly concentrated along the diagonal line, indicating better prediction accuracy from objective model prediction to human evaluation.

\section{Conclusion}

In this work, we attempt to understand SR IQA as a 2D problem in terms of DF and SF. We show that different SR algorithms and categories of SR algorithms exhibit dramatically different behaviors in the 2D space of (DF, SF). We develop separate DF and SF models as well as an uncertainty weighting scheme that combines both fidelity models into an overall quality prediction. The resulting SRIF index outperforms existing methods in the literature when tested using subject-rated SR IQA databases.

In the future, the current work may be extended by developing better DF and SF models or fusion schemes. It would also be interesting to use the 2D approach as a tool to investigate deeper into different SR algorithms to observe how different operations navigate in the 2D space. Furthermore, the 2D approach and the proposed SRIF model may be used to redefine the cost function and improve the design of future SR algorithms for optimal perceptual quality.



\begin{acks}
	This work is in part supported by Natural Sciences and Engineering Research Council of Canada and Canada Research Chair program.
\end{acks}

\bibliographystyle{ACM-Reference-Format}
\bibliography{sample-base}


\begin{thebibliography}{50}


\ifx \showCODEN    \undefined \def \showCODEN     #1{\unskip}     \fi
\ifx \showDOI      \undefined \def \showDOI       #1{#1}\fi
\ifx \showISBNx    \undefined \def \showISBNx     #1{\unskip}     \fi
\ifx \showISBNxiii \undefined \def \showISBNxiii  #1{\unskip}     \fi
\ifx \showISSN     \undefined \def \showISSN      #1{\unskip}     \fi
\ifx \showLCCN     \undefined \def \showLCCN      #1{\unskip}     \fi
\ifx \shownote     \undefined \def \shownote      #1{#1}          \fi
\ifx \showarticletitle \undefined \def \showarticletitle #1{#1}   \fi
\ifx \showURL      \undefined \def \showURL       {\relax}        \fi
\providecommand\bibfield[2]{#2}
\providecommand\bibinfo[2]{#2}
\providecommand\natexlab[1]{#1}
\providecommand\showeprint[2][]{arXiv:#2}

\bibitem[\protect\citeauthoryear{Burt and Adelson}{Burt and Adelson}{1983}]%
        {burt1983laplacian}
\bibfield{author}{\bibinfo{person}{Peter Burt} {and} \bibinfo{person}{Edward
  Adelson}.} \bibinfo{year}{1983}\natexlab{}.
\newblock \showarticletitle{The Laplacian pyramid as a compact image code}.
\newblock \bibinfo{journal}{\emph{IEEE Transactions on Communications}}
  \bibinfo{volume}{31}, \bibinfo{number}{4} (\bibinfo{year}{1983}),
  \bibinfo{pages}{532--540}.
\newblock


\bibitem[\protect\citeauthoryear{Chang, Yeung, and Xiong}{Chang
  et~al\mbox{.}}{2004}]%
        {chang2004super}
\bibfield{author}{\bibinfo{person}{Hong Chang}, \bibinfo{person}{Dit-Yan
  Yeung}, {and} \bibinfo{person}{Yimin Xiong}.}
  \bibinfo{year}{2004}\natexlab{}.
\newblock \showarticletitle{Super-resolution through neighbor embedding}. In
  \bibinfo{booktitle}{\emph{CVPR}}. \bibinfo{pages}{I--I}.
\newblock


\bibitem[\protect\citeauthoryear{Freeman, Jones, and Pasztor}{Freeman
  et~al\mbox{.}}{2002}]%
        {freeman2002example}
\bibfield{author}{\bibinfo{person}{William~T Freeman},
  \bibinfo{person}{Thouis~R Jones}, {and} \bibinfo{person}{Egon~C Pasztor}.}
  \bibinfo{year}{2002}\natexlab{}.
\newblock \showarticletitle{Example-based super-resolution}.
\newblock \bibinfo{journal}{\emph{IEEE Computer Graphics and Applications}}
  \bibinfo{volume}{22}, \bibinfo{number}{2} (\bibinfo{year}{2002}),
  \bibinfo{pages}{56--65}.
\newblock


\bibitem[\protect\citeauthoryear{Guo, Ma, Zhang, Zhou, and Guo}{Guo
  et~al\mbox{.}}{2020}]%
        {guo2020dual}
\bibfield{author}{\bibinfo{person}{Jingcai Guo}, \bibinfo{person}{Shiheng Ma},
  \bibinfo{person}{Jie Zhang}, \bibinfo{person}{Qihua Zhou}, {and}
  \bibinfo{person}{Song Guo}.} \bibinfo{year}{2020}\natexlab{}.
\newblock \showarticletitle{Dual-view attention networks for single image
  super-resolution}. In \bibinfo{booktitle}{\emph{Proceedings of the 28th ACM
  International Conference on Multimedia}}. \bibinfo{pages}{2728--2736}.
\newblock


\bibitem[\protect\citeauthoryear{Hassen, Wang, and Salama}{Hassen
  et~al\mbox{.}}{2013}]%
        {hassen2013image}
\bibfield{author}{\bibinfo{person}{Rania Hassen}, \bibinfo{person}{Zhou Wang},
  {and} \bibinfo{person}{Magdy~MA Salama}.} \bibinfo{year}{2013}\natexlab{}.
\newblock \showarticletitle{Image sharpness assessment based on local phase
  coherence}.
\newblock \bibinfo{journal}{\emph{IEEE Transactions on Image Processing}}
  \bibinfo{volume}{22}, \bibinfo{number}{7} (\bibinfo{year}{2013}),
  \bibinfo{pages}{2798--2810}.
\newblock


\bibitem[\protect\citeauthoryear{Hosseini, Zhang, and Plataniotis}{Hosseini
  et~al\mbox{.}}{2019}]%
        {hosseini2019encoding}
\bibfield{author}{\bibinfo{person}{Mahdi~S Hosseini}, \bibinfo{person}{Yueyang
  Zhang}, {and} \bibinfo{person}{Konstantinos~N Plataniotis}.}
  \bibinfo{year}{2019}\natexlab{}.
\newblock \showarticletitle{Encoding visual sensitivity by maxpol convolution
  filters for image sharpness assessment}.
\newblock \bibinfo{journal}{\emph{IEEE Transactions on Image Processing}}
  \bibinfo{volume}{28}, \bibinfo{number}{9} (\bibinfo{year}{2019}),
  \bibinfo{pages}{4510--4525}.
\newblock


\bibitem[\protect\citeauthoryear{Keys}{Keys}{1981}]%
        {keys1981cubic}
\bibfield{author}{\bibinfo{person}{Robert Keys}.}
  \bibinfo{year}{1981}\natexlab{}.
\newblock \showarticletitle{Cubic convolution interpolation for digital image
  processing}.
\newblock \bibinfo{journal}{\emph{IEEE Transactions on Acoustics, Speech, and
  Signal Processing}} \bibinfo{volume}{29}, \bibinfo{number}{6}
  (\bibinfo{year}{1981}), \bibinfo{pages}{1153--1160}.
\newblock


\bibitem[\protect\citeauthoryear{Kim, Kwon~Lee, and Mu~Lee}{Kim
  et~al\mbox{.}}{2016}]%
        {kim2016accurate}
\bibfield{author}{\bibinfo{person}{Jiwon Kim}, \bibinfo{person}{Jung Kwon~Lee},
  {and} \bibinfo{person}{Kyoung Mu~Lee}.} \bibinfo{year}{2016}\natexlab{}.
\newblock \showarticletitle{Accurate image super-resolution using very deep
  convolutional networks}. In \bibinfo{booktitle}{\emph{CVPR}}.
  \bibinfo{pages}{1646--1654}.
\newblock


\bibitem[\protect\citeauthoryear{Lai, Huang, Ahuja, and Yang}{Lai
  et~al\mbox{.}}{2017}]%
        {lai2017deep}
\bibfield{author}{\bibinfo{person}{Wei-Sheng Lai}, \bibinfo{person}{Jia-Bin
  Huang}, \bibinfo{person}{Narendra Ahuja}, {and} \bibinfo{person}{Ming-Hsuan
  Yang}.} \bibinfo{year}{2017}\natexlab{}.
\newblock \showarticletitle{Deep laplacian pyramid networks for fast and
  accurate super-resolution}. In \bibinfo{booktitle}{\emph{CVPR}}.
  \bibinfo{pages}{624--632}.
\newblock


\bibitem[\protect\citeauthoryear{Ledig, Theis, Husz{\'a}r, Caballero,
  Cunningham, Acosta, Aitken, Tejani, Totz, Wang, et~al\mbox{.}}{Ledig
  et~al\mbox{.}}{2017}]%
        {ledig2017photo}
\bibfield{author}{\bibinfo{person}{Christian Ledig}, \bibinfo{person}{Lucas
  Theis}, \bibinfo{person}{Ferenc Husz{\'a}r}, \bibinfo{person}{Jose
  Caballero}, \bibinfo{person}{Andrew Cunningham}, \bibinfo{person}{Alejandro
  Acosta}, \bibinfo{person}{Andrew Aitken}, \bibinfo{person}{Alykhan Tejani},
  \bibinfo{person}{Johannes Totz}, \bibinfo{person}{Zehan Wang},
  {et~al\mbox{.}}} \bibinfo{year}{2017}\natexlab{}.
\newblock \showarticletitle{Photo-realistic single image super-resolution using
  a generative adversarial network}. In \bibinfo{booktitle}{\emph{CVPR}}.
  \bibinfo{pages}{4681--4690}.
\newblock


\bibitem[\protect\citeauthoryear{Liu, Lin, and Narwaria}{Liu
  et~al\mbox{.}}{2011}]%
        {liu2011image}
\bibfield{author}{\bibinfo{person}{Anmin Liu}, \bibinfo{person}{Weisi Lin},
  {and} \bibinfo{person}{Manish Narwaria}.} \bibinfo{year}{2011}\natexlab{}.
\newblock \showarticletitle{Image quality assessment based on gradient
  similarity}.
\newblock \bibinfo{journal}{\emph{IEEE Transactions on Image Processing}}
  \bibinfo{volume}{21}, \bibinfo{number}{4} (\bibinfo{year}{2011}),
  \bibinfo{pages}{1500--1512}.
\newblock


\bibitem[\protect\citeauthoryear{Ma, Yang, Yang, and Yang}{Ma
  et~al\mbox{.}}{2017c}]%
        {ma2017learning}
\bibfield{author}{\bibinfo{person}{Chao Ma}, \bibinfo{person}{Chih-Yuan Yang},
  \bibinfo{person}{Xiaokang Yang}, {and} \bibinfo{person}{Ming-Hsuan Yang}.}
  \bibinfo{year}{2017}\natexlab{c}.
\newblock \showarticletitle{Learning a no-reference quality metric for
  single-image super-resolution}.
\newblock \bibinfo{journal}{\emph{Computer Vision and Image Understanding}}
  \bibinfo{volume}{158} (\bibinfo{year}{2017}), \bibinfo{pages}{1--16}.
\newblock


\bibitem[\protect\citeauthoryear{Ma, Liu, Liu, Wang, and Tao}{Ma
  et~al\mbox{.}}{2017a}]%
        {ma2017dipiq}
\bibfield{author}{\bibinfo{person}{Kede Ma}, \bibinfo{person}{Wentao Liu},
  \bibinfo{person}{Tongliang Liu}, \bibinfo{person}{Zhou Wang}, {and}
  \bibinfo{person}{Dacheng Tao}.} \bibinfo{year}{2017}\natexlab{a}.
\newblock \showarticletitle{{dipIQ}: Blind image quality assessment by
  learning-to-rank discriminable image pairs}.
\newblock \bibinfo{journal}{\emph{IEEE Transactions on Image Processing}}
  \bibinfo{volume}{26}, \bibinfo{number}{8} (\bibinfo{year}{2017}),
  \bibinfo{pages}{3951--3964}.
\newblock


\bibitem[\protect\citeauthoryear{Ma, Liu, Zhang, Duanmu, Wang, and Zuo}{Ma
  et~al\mbox{.}}{2017b}]%
        {ma2017end}
\bibfield{author}{\bibinfo{person}{Kede Ma}, \bibinfo{person}{Wentao Liu},
  \bibinfo{person}{Kai Zhang}, \bibinfo{person}{Zhengfang Duanmu},
  \bibinfo{person}{Zhou Wang}, {and} \bibinfo{person}{Wangmeng Zuo}.}
  \bibinfo{year}{2017}\natexlab{b}.
\newblock \showarticletitle{End-to-end blind image quality assessment using
  deep neural networks}.
\newblock \bibinfo{journal}{\emph{IEEE Transactions on Image Processing}}
  \bibinfo{volume}{27}, \bibinfo{number}{3} (\bibinfo{year}{2017}),
  \bibinfo{pages}{1202--1213}.
\newblock


\bibitem[\protect\citeauthoryear{Mittal, Moorthy, and Bovik}{Mittal
  et~al\mbox{.}}{2012a}]%
        {mittal2012no}
\bibfield{author}{\bibinfo{person}{Anish Mittal},
  \bibinfo{person}{Anush~Krishna Moorthy}, {and} \bibinfo{person}{Alan~Conrad
  Bovik}.} \bibinfo{year}{2012}\natexlab{a}.
\newblock \showarticletitle{No-reference image quality assessment in the
  spatial domain}.
\newblock \bibinfo{journal}{\emph{IEEE Transactions on Image Processing}}
  \bibinfo{volume}{21}, \bibinfo{number}{12} (\bibinfo{year}{2012}),
  \bibinfo{pages}{4695--4708}.
\newblock


\bibitem[\protect\citeauthoryear{Mittal, Soundararajan, and Bovik}{Mittal
  et~al\mbox{.}}{2012b}]%
        {mittal2012making}
\bibfield{author}{\bibinfo{person}{Anish Mittal}, \bibinfo{person}{Rajiv
  Soundararajan}, {and} \bibinfo{person}{Alan~C Bovik}.}
  \bibinfo{year}{2012}\natexlab{b}.
\newblock \showarticletitle{Making a ``completely blind'' image quality
  analyzer}.
\newblock \bibinfo{journal}{\emph{IEEE Signal Processing Letters}}
  \bibinfo{volume}{20}, \bibinfo{number}{3} (\bibinfo{year}{2012}),
  \bibinfo{pages}{209--212}.
\newblock


\bibitem[\protect\citeauthoryear{Moorthy and Bovik}{Moorthy and Bovik}{2011}]%
        {moorthy2011blind}
\bibfield{author}{\bibinfo{person}{Anush~Krishna Moorthy} {and}
  \bibinfo{person}{Alan~Conrad Bovik}.} \bibinfo{year}{2011}\natexlab{}.
\newblock \showarticletitle{Blind image quality assessment: From natural scene
  statistics to perceptual quality}.
\newblock \bibinfo{journal}{\emph{IEEE Transactions on Image Processing}}
  \bibinfo{volume}{20}, \bibinfo{number}{12} (\bibinfo{year}{2011}),
  \bibinfo{pages}{3350--3364}.
\newblock


\bibitem[\protect\citeauthoryear{Park, Park, and Kang}{Park
  et~al\mbox{.}}{2003}]%
        {park2003super}
\bibfield{author}{\bibinfo{person}{Sung~Cheol Park}, \bibinfo{person}{Min~Kyu
  Park}, {and} \bibinfo{person}{Moon~Gi Kang}.}
  \bibinfo{year}{2003}\natexlab{}.
\newblock \showarticletitle{Super-resolution image reconstruction: a technical
  overview}.
\newblock \bibinfo{journal}{\emph{IEEE Signal Processing Magazine}}
  \bibinfo{volume}{20}, \bibinfo{number}{3} (\bibinfo{year}{2003}),
  \bibinfo{pages}{21--36}.
\newblock


\bibitem[\protect\citeauthoryear{Reibman, Bell, and Gray}{Reibman
  et~al\mbox{.}}{2006}]%
        {reibman2006quality}
\bibfield{author}{\bibinfo{person}{Amy~R Reibman}, \bibinfo{person}{Robert~M
  Bell}, {and} \bibinfo{person}{Sharon Gray}.} \bibinfo{year}{2006}\natexlab{}.
\newblock \showarticletitle{Quality assessment for super-resolution image
  enhancement}. In \bibinfo{booktitle}{\emph{IEEE International Conference on
  Image Processing}}. \bibinfo{pages}{2017--2020}.
\newblock


\bibitem[\protect\citeauthoryear{Rohaly, Corriveau, Libert, Webster, Baroncini,
  Beerends, Blin, Contin, Hamada, Harrison, et~al\mbox{.}}{Rohaly
  et~al\mbox{.}}{2000}]%
        {rohaly2000video}
\bibfield{author}{\bibinfo{person}{Ann~Marie Rohaly}, \bibinfo{person}{Philip~J
  Corriveau}, \bibinfo{person}{John~M Libert}, \bibinfo{person}{Arthur~A
  Webster}, \bibinfo{person}{Vittorio Baroncini}, \bibinfo{person}{John
  Beerends}, \bibinfo{person}{Jean-Louis Blin}, \bibinfo{person}{Laura Contin},
  \bibinfo{person}{Takahiro Hamada}, \bibinfo{person}{David Harrison},
  {et~al\mbox{.}}} \bibinfo{year}{2000}\natexlab{}.
\newblock \showarticletitle{Video quality experts group: Current results and
  future directions}. In \bibinfo{booktitle}{\emph{Visual Communications and
  Image Processing}}, Vol.~\bibinfo{volume}{4067}. SPIE,
  \bibinfo{pages}{742--753}.
\newblock


\bibitem[\protect\citeauthoryear{Saad, Bovik, and Charrier}{Saad
  et~al\mbox{.}}{2012}]%
        {saad2012blind}
\bibfield{author}{\bibinfo{person}{Michele~A Saad}, \bibinfo{person}{Alan~C
  Bovik}, {and} \bibinfo{person}{Christophe Charrier}.}
  \bibinfo{year}{2012}\natexlab{}.
\newblock \showarticletitle{Blind image quality assessment: A natural scene
  statistics approach in the DCT domain}.
\newblock \bibinfo{journal}{\emph{IEEE Transactions on Image Processing}}
  \bibinfo{volume}{21}, \bibinfo{number}{8} (\bibinfo{year}{2012}),
  \bibinfo{pages}{3339--3352}.
\newblock


\bibitem[\protect\citeauthoryear{Sampat, Wang, Gupta, Bovik, and Markey}{Sampat
  et~al\mbox{.}}{2009}]%
        {sampat2009complex}
\bibfield{author}{\bibinfo{person}{Mehul~P Sampat}, \bibinfo{person}{Zhou
  Wang}, \bibinfo{person}{Shalini Gupta}, \bibinfo{person}{Alan~Conrad Bovik},
  {and} \bibinfo{person}{Mia~K Markey}.} \bibinfo{year}{2009}\natexlab{}.
\newblock \showarticletitle{Complex wavelet structural similarity: A new image
  similarity index}.
\newblock \bibinfo{journal}{\emph{IEEE Transactions on Image Processing}}
  \bibinfo{volume}{18}, \bibinfo{number}{11} (\bibinfo{year}{2009}),
  \bibinfo{pages}{2385--2401}.
\newblock


\bibitem[\protect\citeauthoryear{Sheikh and Bovik}{Sheikh and Bovik}{2006}]%
        {sheikh2006image}
\bibfield{author}{\bibinfo{person}{Hamid~R Sheikh} {and}
  \bibinfo{person}{Alan~C Bovik}.} \bibinfo{year}{2006}\natexlab{}.
\newblock \showarticletitle{Image information and visual quality}.
\newblock \bibinfo{journal}{\emph{IEEE Transactions on Image Processing}}
  \bibinfo{volume}{15}, \bibinfo{number}{2} (\bibinfo{year}{2006}),
  \bibinfo{pages}{430--444}.
\newblock


\bibitem[\protect\citeauthoryear{Sheikh, Bovik, and De~Veciana}{Sheikh
  et~al\mbox{.}}{2005}]%
        {sheikh2005information}
\bibfield{author}{\bibinfo{person}{Hamid~R Sheikh}, \bibinfo{person}{Alan~C
  Bovik}, {and} \bibinfo{person}{Gustavo De~Veciana}.}
  \bibinfo{year}{2005}\natexlab{}.
\newblock \showarticletitle{An information fidelity criterion for image quality
  assessment using natural scene statistics}.
\newblock \bibinfo{journal}{\emph{IEEE Transactions on Image Processing}}
  \bibinfo{volume}{14}, \bibinfo{number}{12} (\bibinfo{year}{2005}),
  \bibinfo{pages}{2117--2128}.
\newblock


\bibitem[\protect\citeauthoryear{Shi, Wan, Wu, Xie, Dong, and Wu}{Shi
  et~al\mbox{.}}{2019}]%
        {shi2019sisrset}
\bibfield{author}{\bibinfo{person}{Guangming Shi}, \bibinfo{person}{Wenfei
  Wan}, \bibinfo{person}{Jinjian Wu}, \bibinfo{person}{Xuemei Xie},
  \bibinfo{person}{Weisheng Dong}, {and} \bibinfo{person}{Hong~Ren Wu}.}
  \bibinfo{year}{2019}\natexlab{}.
\newblock \showarticletitle{SISRSet: Single image super-resolution subjective
  evaluation test and objective quality assessment}.
\newblock \bibinfo{journal}{\emph{Neurocomputing}}  \bibinfo{volume}{360}
  (\bibinfo{year}{2019}), \bibinfo{pages}{37--51}.
\newblock


\bibitem[\protect\citeauthoryear{Simonyan and Zisserman}{Simonyan and
  Zisserman}{2014}]%
        {simonyan2014very}
\bibfield{author}{\bibinfo{person}{Karen Simonyan} {and}
  \bibinfo{person}{Andrew Zisserman}.} \bibinfo{year}{2014}\natexlab{}.
\newblock \showarticletitle{Very deep convolutional networks for large-scale
  image recognition}.
\newblock \bibinfo{journal}{\emph{arXiv preprint arXiv:1409.1556}}
  (\bibinfo{year}{2014}).
\newblock


\bibitem[\protect\citeauthoryear{Sun, Liao, Xue, and Zhou}{Sun
  et~al\mbox{.}}{2018}]%
        {sun2018spsim}
\bibfield{author}{\bibinfo{person}{Wen Sun}, \bibinfo{person}{Qingmin Liao},
  \bibinfo{person}{Jing-Hao Xue}, {and} \bibinfo{person}{Fei Zhou}.}
  \bibinfo{year}{2018}\natexlab{}.
\newblock \showarticletitle{{SPSIM}: A superpixel-based similarity index for
  full-reference image quality assessment}.
\newblock \bibinfo{journal}{\emph{IEEE Transactions on Image Processing}}
  \bibinfo{volume}{27}, \bibinfo{number}{9} (\bibinfo{year}{2018}),
  \bibinfo{pages}{4232--4244}.
\newblock


\bibitem[\protect\citeauthoryear{Tai, Yang, and Liu}{Tai et~al\mbox{.}}{2017}]%
        {tai2017image}
\bibfield{author}{\bibinfo{person}{Ying Tai}, \bibinfo{person}{Jian Yang},
  {and} \bibinfo{person}{Xiaoming Liu}.} \bibinfo{year}{2017}\natexlab{}.
\newblock \showarticletitle{Image super-resolution via deep recursive residual
  network}. In \bibinfo{booktitle}{\emph{CVPR}}. \bibinfo{pages}{3147--3155}.
\newblock


\bibitem[\protect\citeauthoryear{Vu, Phan, and Chandler}{Vu
  et~al\mbox{.}}{2011}]%
        {vu2011bf}
\bibfield{author}{\bibinfo{person}{Cuong~T Vu}, \bibinfo{person}{Thien~D Phan},
  {and} \bibinfo{person}{Damon~M Chandler}.} \bibinfo{year}{2011}\natexlab{}.
\newblock \showarticletitle{{S3}: A spectral and spatial measure of local
  perceived sharpness in natural images}.
\newblock \bibinfo{journal}{\emph{IEEE Transactions on Image Processing}}
  \bibinfo{volume}{21}, \bibinfo{number}{3} (\bibinfo{year}{2011}),
  \bibinfo{pages}{934--945}.
\newblock


\bibitem[\protect\citeauthoryear{Wang, Li, Li, Gu, Lu, and Qian}{Wang
  et~al\mbox{.}}{2017}]%
        {wang2017perceptual}
\bibfield{author}{\bibinfo{person}{Guangcheng Wang}, \bibinfo{person}{Leida
  Li}, \bibinfo{person}{Qiaohong Li}, \bibinfo{person}{Ke Gu},
  \bibinfo{person}{Zhaolin Lu}, {and} \bibinfo{person}{Jiansheng Qian}.}
  \bibinfo{year}{2017}\natexlab{}.
\newblock \showarticletitle{Perceptual evaluation of single-image
  super-resolution reconstruction}. In \bibinfo{booktitle}{\emph{IEEE
  International Conference on Image Processing}}. \bibinfo{pages}{3145--3149}.
\newblock


\bibitem[\protect\citeauthoryear{Wang and Ward}{Wang and Ward}{2007}]%
        {wang2007new}
\bibfield{author}{\bibinfo{person}{Qing Wang} {and}
  \bibinfo{person}{Rabab~Kreidieh Ward}.} \bibinfo{year}{2007}\natexlab{}.
\newblock \showarticletitle{A new orientation-adaptive interpolation method}.
\newblock \bibinfo{journal}{\emph{IEEE Transactions on Image Processing}}
  \bibinfo{volume}{16}, \bibinfo{number}{4} (\bibinfo{year}{2007}),
  \bibinfo{pages}{889--900}.
\newblock


\bibitem[\protect\citeauthoryear{Wang, Zhang, Liang, and Pan}{Wang
  et~al\mbox{.}}{2012}]%
        {wang2012semi}
\bibfield{author}{\bibinfo{person}{Shenlong Wang}, \bibinfo{person}{Lei Zhang},
  \bibinfo{person}{Yan Liang}, {and} \bibinfo{person}{Quan Pan}.}
  \bibinfo{year}{2012}\natexlab{}.
\newblock \showarticletitle{Semi-coupled dictionary learning with applications
  to image super-resolution and photo-sketch synthesis}. In
  \bibinfo{booktitle}{\emph{CVPR}}. \bibinfo{pages}{2216--2223}.
\newblock


\bibitem[\protect\citeauthoryear{Wang, Bovik, Sheikh, and Simoncelli}{Wang
  et~al\mbox{.}}{2004}]%
        {wang2004image}
\bibfield{author}{\bibinfo{person}{Zhou Wang}, \bibinfo{person}{Alan~C Bovik},
  \bibinfo{person}{Hamid~R Sheikh}, {and} \bibinfo{person}{Eero~P Simoncelli}.}
  \bibinfo{year}{2004}\natexlab{}.
\newblock \showarticletitle{Image quality assessment: from error visibility to
  structural similarity}.
\newblock \bibinfo{journal}{\emph{IEEE Transactions on Image Processing}}
  \bibinfo{volume}{13}, \bibinfo{number}{4} (\bibinfo{year}{2004}),
  \bibinfo{pages}{600--612}.
\newblock


\bibitem[\protect\citeauthoryear{Wang, Chen, and Hoi}{Wang
  et~al\mbox{.}}{2020}]%
        {wang2020deep}
\bibfield{author}{\bibinfo{person}{Zhihao Wang}, \bibinfo{person}{Jian Chen},
  {and} \bibinfo{person}{Steven~CH Hoi}.} \bibinfo{year}{2020}\natexlab{}.
\newblock \showarticletitle{Deep learning for image super-resolution: A
  survey}.
\newblock \bibinfo{journal}{\emph{IEEE Transactions on Pattern Analysis and
  Machine Intelligence}} \bibinfo{volume}{43}, \bibinfo{number}{10}
  (\bibinfo{year}{2020}), \bibinfo{pages}{3365--3387}.
\newblock


\bibitem[\protect\citeauthoryear{Wang and Li}{Wang and Li}{2010}]%
        {wang2010information}
\bibfield{author}{\bibinfo{person}{Zhou Wang} {and} \bibinfo{person}{Qiang
  Li}.} \bibinfo{year}{2010}\natexlab{}.
\newblock \showarticletitle{Information content weighting for perceptual image
  quality assessment}.
\newblock \bibinfo{journal}{\emph{IEEE Transactions on Image Processing}}
  \bibinfo{volume}{20}, \bibinfo{number}{5} (\bibinfo{year}{2010}),
  \bibinfo{pages}{1185--1198}.
\newblock


\bibitem[\protect\citeauthoryear{Wang, Simoncelli, and Bovik}{Wang
  et~al\mbox{.}}{2003}]%
        {wang2003multiscale}
\bibfield{author}{\bibinfo{person}{Zhou Wang}, \bibinfo{person}{Eero~P
  Simoncelli}, {and} \bibinfo{person}{Alan~C Bovik}.}
  \bibinfo{year}{2003}\natexlab{}.
\newblock \showarticletitle{Multiscale structural similarity for image quality
  assessment}. In \bibinfo{booktitle}{\emph{The Thrity-Seventh Asilomar
  Conference on Signals, Systems \& Computers, 2003}},
  Vol.~\bibinfo{volume}{2}. IEEE, \bibinfo{pages}{1398--1402}.
\newblock


\bibitem[\protect\citeauthoryear{Wu, Lin, Shi, and Liu}{Wu
  et~al\mbox{.}}{2012}]%
        {wu2012perceptual}
\bibfield{author}{\bibinfo{person}{Jinjian Wu}, \bibinfo{person}{Weisi Lin},
  \bibinfo{person}{Guangming Shi}, {and} \bibinfo{person}{Anmin Liu}.}
  \bibinfo{year}{2012}\natexlab{}.
\newblock \showarticletitle{Perceptual quality metric with internal generative
  mechanism}.
\newblock \bibinfo{journal}{\emph{IEEE Transactions on Image Processing}}
  \bibinfo{volume}{22}, \bibinfo{number}{1} (\bibinfo{year}{2012}),
  \bibinfo{pages}{43--54}.
\newblock


\bibitem[\protect\citeauthoryear{Wu, Wang, and Li}{Wu et~al\mbox{.}}{2015}]%
        {wu2015highly}
\bibfield{author}{\bibinfo{person}{Qingbo Wu}, \bibinfo{person}{Zhou Wang},
  {and} \bibinfo{person}{Hongliang Li}.} \bibinfo{year}{2015}\natexlab{}.
\newblock \showarticletitle{A highly efficient method for blind image quality
  assessment}. In \bibinfo{booktitle}{\emph{IEEE International Conference on
  Image Processing}}. \bibinfo{pages}{339--343}.
\newblock


\bibitem[\protect\citeauthoryear{Xiao, Ye, Zhao, Lam, and Wan}{Xiao
  et~al\mbox{.}}{2021}]%
        {xiao2021self}
\bibfield{author}{\bibinfo{person}{Jun Xiao}, \bibinfo{person}{Qian Ye},
  \bibinfo{person}{Rui Zhao}, \bibinfo{person}{Kin-Man Lam}, {and}
  \bibinfo{person}{Kao Wan}.} \bibinfo{year}{2021}\natexlab{}.
\newblock \showarticletitle{Self-feature Learning: An Efficient Deep
  Lightweight Network for Image Super-resolution}. In
  \bibinfo{booktitle}{\emph{Proceedings of the 29th ACM International
  Conference on Multimedia}}. \bibinfo{pages}{4408--4416}.
\newblock


\bibitem[\protect\citeauthoryear{Xue, Zhang, Mou, and Bovik}{Xue
  et~al\mbox{.}}{2013}]%
        {xue2013gradient}
\bibfield{author}{\bibinfo{person}{Wufeng Xue}, \bibinfo{person}{Lei Zhang},
  \bibinfo{person}{Xuanqin Mou}, {and} \bibinfo{person}{Alan~C Bovik}.}
  \bibinfo{year}{2013}\natexlab{}.
\newblock \showarticletitle{Gradient magnitude similarity deviation: A highly
  efficient perceptual image quality index}.
\newblock \bibinfo{journal}{\emph{IEEE Transactions on Image Processing}}
  \bibinfo{volume}{23}, \bibinfo{number}{2} (\bibinfo{year}{2013}),
  \bibinfo{pages}{684--695}.
\newblock


\bibitem[\protect\citeauthoryear{Yamanaka, Kuwashima, and Kurita}{Yamanaka
  et~al\mbox{.}}{2017}]%
        {yamanaka2017fast}
\bibfield{author}{\bibinfo{person}{Jin Yamanaka}, \bibinfo{person}{Shigesumi
  Kuwashima}, {and} \bibinfo{person}{Takio Kurita}.}
  \bibinfo{year}{2017}\natexlab{}.
\newblock \showarticletitle{Fast and accurate image super resolution by deep
  CNN with skip connection and network in network}. In
  \bibinfo{booktitle}{\emph{International Conference on Neural Information
  Processing}}. Springer, \bibinfo{pages}{217--225}.
\newblock


\bibitem[\protect\citeauthoryear{Yang, Ma, and Yang}{Yang
  et~al\mbox{.}}{2014}]%
        {yang2014single}
\bibfield{author}{\bibinfo{person}{Chih-Yuan Yang}, \bibinfo{person}{Chao Ma},
  {and} \bibinfo{person}{Ming-Hsuan Yang}.} \bibinfo{year}{2014}\natexlab{}.
\newblock \showarticletitle{Single-image super-resolution: A benchmark}. In
  \bibinfo{booktitle}{\emph{ECCV}}. \bibinfo{pages}{372--386}.
\newblock


\bibitem[\protect\citeauthoryear{Yang and Yang}{Yang and Yang}{2013}]%
        {yang2013fast}
\bibfield{author}{\bibinfo{person}{Chih-Yuan Yang} {and}
  \bibinfo{person}{Ming-Hsuan Yang}.} \bibinfo{year}{2013}\natexlab{}.
\newblock \showarticletitle{Fast direct super-resolution by simple functions}.
  In \bibinfo{booktitle}{\emph{ICCV}}. \bibinfo{pages}{561--568}.
\newblock


\bibitem[\protect\citeauthoryear{Yang, Wright, Huang, and Ma}{Yang
  et~al\mbox{.}}{2010}]%
        {yang2010image}
\bibfield{author}{\bibinfo{person}{Jianchao Yang}, \bibinfo{person}{John
  Wright}, \bibinfo{person}{Thomas~S Huang}, {and} \bibinfo{person}{Yi Ma}.}
  \bibinfo{year}{2010}\natexlab{}.
\newblock \showarticletitle{Image super-resolution via sparse representation}.
\newblock \bibinfo{journal}{\emph{IEEE Transactions on Image Processing}}
  \bibinfo{volume}{19}, \bibinfo{number}{11} (\bibinfo{year}{2010}),
  \bibinfo{pages}{2861--2873}.
\newblock


\bibitem[\protect\citeauthoryear{Yang, Tian, Zhou, Liao, Chen, and Zheng}{Yang
  et~al\mbox{.}}{2016}]%
        {yang2016consistent}
\bibfield{author}{\bibinfo{person}{Wenming Yang}, \bibinfo{person}{Yapeng
  Tian}, \bibinfo{person}{Fei Zhou}, \bibinfo{person}{Qingmin Liao},
  \bibinfo{person}{Hai Chen}, {and} \bibinfo{person}{Chenglin Zheng}.}
  \bibinfo{year}{2016}\natexlab{}.
\newblock \showarticletitle{Consistent coding scheme for single-image
  super-resolution via independent dictionaries}.
\newblock \bibinfo{journal}{\emph{IEEE Transactions on Multimedia}}
  \bibinfo{volume}{18}, \bibinfo{number}{3} (\bibinfo{year}{2016}),
  \bibinfo{pages}{313--325}.
\newblock


\bibitem[\protect\citeauthoryear{Yeganeh, Rostami, and Wang}{Yeganeh
  et~al\mbox{.}}{2015}]%
        {yeganeh2015objective}
\bibfield{author}{\bibinfo{person}{Hojatollah Yeganeh},
  \bibinfo{person}{Mohammad Rostami}, {and} \bibinfo{person}{Zhou Wang}.}
  \bibinfo{year}{2015}\natexlab{}.
\newblock \showarticletitle{Objective quality assessment of interpolated
  natural images}.
\newblock \bibinfo{journal}{\emph{IEEE Transactions on Image Processing}}
  \bibinfo{volume}{24}, \bibinfo{number}{11} (\bibinfo{year}{2015}),
  \bibinfo{pages}{4651--4663}.
\newblock


\bibitem[\protect\citeauthoryear{Ying, Niu, Gupta, Mahajan, Ghadiyaram, and
  Bovik}{Ying et~al\mbox{.}}{2020}]%
        {ying2020patches}
\bibfield{author}{\bibinfo{person}{Zhenqiang Ying}, \bibinfo{person}{Haoran
  Niu}, \bibinfo{person}{Praful Gupta}, \bibinfo{person}{Dhruv Mahajan},
  \bibinfo{person}{Deepti Ghadiyaram}, {and} \bibinfo{person}{Alan Bovik}.}
  \bibinfo{year}{2020}\natexlab{}.
\newblock \showarticletitle{From patches to pictures ({PaQ-2-PiQ}): Mapping the
  perceptual space of picture quality}. In \bibinfo{booktitle}{\emph{CVPR}}.
  \bibinfo{pages}{3575--3585}.
\newblock


\bibitem[\protect\citeauthoryear{Zhang, Zhang, Mou, and Zhang}{Zhang
  et~al\mbox{.}}{2011}]%
        {zhang2011fsim}
\bibfield{author}{\bibinfo{person}{Lin Zhang}, \bibinfo{person}{Lei Zhang},
  \bibinfo{person}{Xuanqin Mou}, {and} \bibinfo{person}{David Zhang}.}
  \bibinfo{year}{2011}\natexlab{}.
\newblock \showarticletitle{{FSIM}: A feature similarity index for image
  quality assessment}.
\newblock \bibinfo{journal}{\emph{IEEE Transactions on Image Processing}}
  \bibinfo{volume}{20}, \bibinfo{number}{8} (\bibinfo{year}{2011}),
  \bibinfo{pages}{2378--2386}.
\newblock


\bibitem[\protect\citeauthoryear{Zhou, Yao, Liu, and Qiu}{Zhou
  et~al\mbox{.}}{2019}]%
        {zhou2019visual}
\bibfield{author}{\bibinfo{person}{Fei Zhou}, \bibinfo{person}{Rongguo Yao},
  \bibinfo{person}{Bozhi Liu}, {and} \bibinfo{person}{Guoping Qiu}.}
  \bibinfo{year}{2019}\natexlab{}.
\newblock \showarticletitle{Visual quality assessment for super-resolved
  images: database and method}.
\newblock \bibinfo{journal}{\emph{IEEE Transactions on Image Processing}}
  \bibinfo{volume}{28}, \bibinfo{number}{7} (\bibinfo{year}{2019}),
  \bibinfo{pages}{3528--3541}.
\newblock


\bibitem[\protect\citeauthoryear{Zhou, Wang, and Chen}{Zhou
  et~al\mbox{.}}{2021}]%
        {zhou2021image}
\bibfield{author}{\bibinfo{person}{Wei Zhou}, \bibinfo{person}{Zhou Wang},
  {and} \bibinfo{person}{Zhibo Chen}.} \bibinfo{year}{2021}\natexlab{}.
\newblock \showarticletitle{Image super-resolution quality assessment:
  Structural fidelity versus statistical naturalness}. In
  \bibinfo{booktitle}{\emph{IEEE International Conference on Quality of
  Multimedia Experience}}. \bibinfo{pages}{61--64}.
\newblock


\end{thebibliography}


\end{document}